\pdfoutput=1
\documentclass[11pt]{article}

\usepackage[preprint]{acl}

\usepackage{times}
\usepackage{latexsym}
\usepackage{multirow}

\usepackage[T1]{fontenc}
\usepackage[utf8]{inputenc}

\usepackage{microtype}

\usepackage{inconsolata}

\usepackage{graphicx}
\usepackage{booktabs}
\usepackage{amsmath,amsthm,amssymb}
\usepackage{mathrsfs}

\usepackage[most,skins,theorems]{tcolorbox}
\tcbset{
  aibox/.style={
    width=\linewidth,
    top=10pt,
    bottom=4pt,
    colback=blue!6!white,
    colframe=black,
    colbacktitle=black,
    enhanced,
    center,
    attach boxed title to top left={yshift=-0.1in,xshift=0.15in},
    boxed title style={boxrule=0pt,colframe=white,},
  }
}

\definecolor{rliableolive}{HTML}{BBCC33}
\definecolor{rliableblue}{HTML}{77AADD}
\definecolor{rliablered}{HTML}{EE8866}

\newtcolorbox{questionlayer}[1][]{
  colback=blue!10!white, colframe=blue!70, sharp corners, boxrule=0.5pt, % Blue layer for questions
  title=Question: #1
}

\newtcolorbox{answerlayer}[1][]{
  colback=green!5!white, colframe=green!75!black, sharp corners, boxrule=0.5pt, % Green layer for answers
  title=Answer:
}

\newtcolorbox{reasonlayer}[1][]{
  colback=yellow!10!white, colframe=yellow!75!black, sharp corners, boxrule=0.5pt, % Yellow layer for bot responses
  title=Reason:
}
\newtcolorbox{multiqa}[1][]{
  colback=gray!10, colframe=black, sharp corners, boxrule=1pt, % Outer container box
  fonttitle=\bfseries,
  title=Q\&A Case, % Display the counter value
}

\newtcolorbox{AIbox}[2][]{aibox,title=#2,colback=rliableblue!10!white,#1}

\newcommand{\wyd}[1]{\textcolor{black}{#1}} 

\title{From Mathematical Reasoning to Code: Generalization of Process Reward Models in Test-Time Scaling}
\author{
 \textbf{Zhengyu Chen\textsuperscript{1}\footnotemark[1]},
 Yudong Wang \textsuperscript{2}\footnotemark[1],
 Teng Xiao\textbf{\textsuperscript{3}},
 Ruochen Zhou \textsuperscript{1},\\
 \textbf{Xuesheng Yang \textsuperscript{4}},
 \textbf{Wei Wang\textsuperscript{1}},
 \textbf{Zhifang Sui \textsuperscript{2}},
 \textbf{Jingang Wang\textsuperscript{1}}
\\
\\
 \textsuperscript{1}Meituan Inc. \\
 \textsuperscript{2} National Key Laboratory for Multimedia Information Processing, Peking University
 \\
 \textsuperscript{3} Pennsylvania State University \textsuperscript{4} Peking University\\
\\
\quad\texttt{\{chenzhengyu04,wangjingang02\}@meituan.com}\\
}
\begin{document}
\maketitle
\renewcommand{\thefootnote}{\fnsymbol{footnote}}
\footnotetext[1]{Equal Contribution.}

\begin{abstract}
% In recent years, Process Reward Models (PRMs) have emerged as a promising approach to enhancing the mathematical reasoning capabilities of Large Language Models by systematically reducing intermediate errors. This study extends the application of PRMs to code generation tasks, exploring their scalability, generalization, and performance under various conditions. We investigate the interplay between pre-training and reward model training FLOPs, analyzing how these computational resources affect PRM efficiency and accuracy in complex reasoning tasks. Our findings reveal diminishing returns in model performance with increasing PRM size, underscoring the importance of balancing model size and computational cost. The diversity of training datasets is shown to significantly influence PRM performance, emphasizing the need for varied data to enhance accuracy and efficiency. Additionally, we explore test-time scaling strategies, identifying Monte Carlo Tree Search (MCTS) as the most effective strategy when resources are abundant, while Best-of-N Sampling provides a practical solution under constraints. Our research demonstrates that PRMs trained on mathematical data perform comparably to those optimized for code generation, indicating promising cross-domain adaptability.

Recent advancements in improving the reasoning capabilities of Large Language Models have underscored the efficacy of Process Reward Models (PRMs) in addressing intermediate errors through structured feedback mechanisms. This study analyzes PRMs from multiple perspectives, including training methodologies, scalability, and generalization capabilities. We investigate the interplay between pre-training and reward model training FLOPs to assess their influence on PRM efficiency and accuracy in complex reasoning tasks. Our analysis reveals a pattern of diminishing returns in performance with increasing PRM scale, highlighting the importance of balancing model size and computational cost. Furthermore, the diversity of training datasets significantly impacts PRM performance, emphasizing the importance of diverse data to enhance both accuracy and efficiency. We further examine test-time scaling strategies, identifying Monte Carlo Tree Search as the most effective method when computational resources are abundant, while Best-of-N Sampling serves as a practical alternative under resource-limited conditions. Notably, our findings indicate that PRMs trained on mathematical datasets exhibit performance comparable to those tailored for code generation, suggesting robust cross-domain generalization. Employing a gradient-based metric, we observe that PRMs exhibit a preference for selecting responses with similar underlying patterns, further informing their optimization.

% This work contributes to the understanding of PRM generalization, offering insights for the development of more adaptive and efficient models across diverse applications.

% In recent years, there have been significant advancements in Large Language Models (LLMs), especially in enhancing their capabilities for mathematical reasoning through the development of Process Reward Models (PRMs). These models are designed to systematically identify and rectify intermediate errors in reasoning processes. While existing research has shown promising results within the domain of mathematics, this study seeks to explore the generalization ability of PRMs, particularly their applicability to code generation and their test-time scaling performance. PRMs trained on mathematical data exhibit remarkable generalization and identification capabilities on previously unseen code generation. The findings reveal that methodologies effective in mathematical reasoning can also succeed in code generation, thereby highlighting PRMs' robustness and effectiveness.
\end{abstract}

% Recent advancements in enhancing the reasoning of Large Language Models have highlighted the potential of Process Reward Models (PRMs) in mitigating intermediate errors through systematic feedback. This work analyzes the characteristics of PRMs from several distinct perspectives, including training computes, scalability, and generalization abilities. We analyze the relationship between pre-training and reward model training FLOPs to determine their impact on PRM efficiency and accuracy in complex reasoning tasks. Our findings indicate a trend of diminishing performance gains with increasing PRM size, emphasizing the critical consideration of balancing model scale and computational expenditure. Furthermore, the heterogeneity of training datasets is shown to exert a significant influence on PRM performance, underscoring the necessity of diverse data for improved accuracy and efficiency. We also explore test-time scaling strategies, identifying Monte Carlo Tree Search (MCTS) as the most effective approach when computational resources are ample, while Best-of-N Sampling offers a viable alternative under resource constraints. Notably, our study demonstrates that PRMs trained on mathematical data achieve comparable performance to those specifically optimized for code generation, suggesting a promising degree of cross-domain adaptability. With a gradient-based metric, we find PRMs prefer to select responses with specific patterns.

\section{Introduction}

In recent years, the advancement of Process Reward Models (PRMs) has garnered significant attention due to their potential to enhance Large Language Models' mathematical reasoning capabilities \cite{zhang2025lessons, lightman2023lets}. These models aim to systematically identify and reduce intermediate errors in reasoning processes \cite{zhang2024entropy, wang2024math}. This targeted approach has yielded promising results in mathematics, leading researchers to explore whether such capabilities extend to other fields, such as code generation. This study extends the exploration of PRMs to code generation tasks, examining their generalization ability and performance under various scaling conditions to optimize PRM training.

Our research investigates the scalability of PRMs by analyzing the interplay between pre-training and reward model training FLOPs. We focus on how these computational resources impact the efficiency and accuracy of PRMs in executing complex reasoning tasks. By evaluating PRMs primarily trained on mathematical data, we aim to assess their robustness in handling novel tasks within the domain of code generation.

% \begin{figure*}[h!]
%     \centering
%     \vspace{-0.15in}
%     \begin{minipage}{0.4\linewidth} 
%         \centering
%         \includegraphics[width=\linewidth]{figure/N_8.pdf}
%         \label{fig:training_loss_tokens_lr_v3}
%     \end{minipage} 
%     % \hfill
%     \begin{minipage}{0.4\linewidth} 
%         \centering
%         \includegraphics[width=\linewidth]{figure/2N_8.pdf}
%         \label{fig:training_loss_tokens_lr_v3}
%     \end{minipage}  
%     % \hfill
%     % \begin{minipage}{0.24\linewidth} 
%     %     \centering
%     %     \includegraphics[width=\linewidth]{figure/N_16.pdf}
%     %      \label{fig:training_loss_steps_1B}
%     % \end{minipage}  
%     % \hfill
%     % \begin{minipage}{0.24\linewidth} 
%     %     \centering
%     %     \includegraphics[width=\linewidth]{figure/2N_16.pdf}
%     %      \label{fig:training_loss_steps_1B}
%     % \end{minipage}  
%     \vspace{-0.25in}
%     \caption{Comparison of the performance of PRMs with varying model sizes during inference using the same language model. As model size increases, accuracy improves rapidly, indicating a better capability to capture the complexity necessary for solving mathematical problems.}
%     \label{fig:PRM_Models}
% \end{figure*}

\begin{figure*}[h!]
    \centering
    \vspace{-0.15in}
    \begin{minipage}{0.4\linewidth} 
        \centering
        \includegraphics[width=\linewidth]{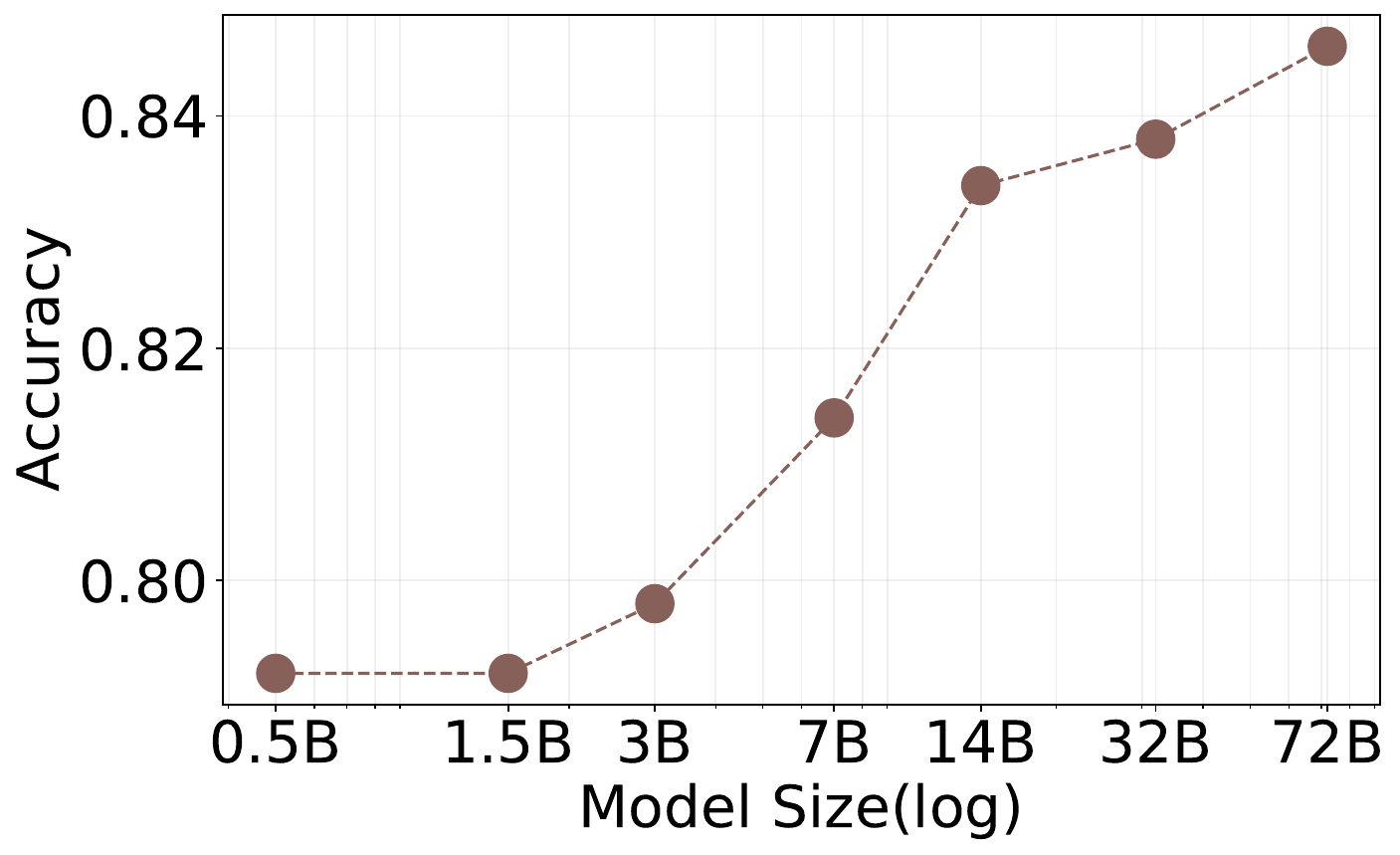}
        \label{fig:training_loss_tokens_lr_v3}
    \end{minipage} 
    % \hfill
    \begin{minipage}{0.4\linewidth} 
        \centering
        \includegraphics[width=\linewidth]{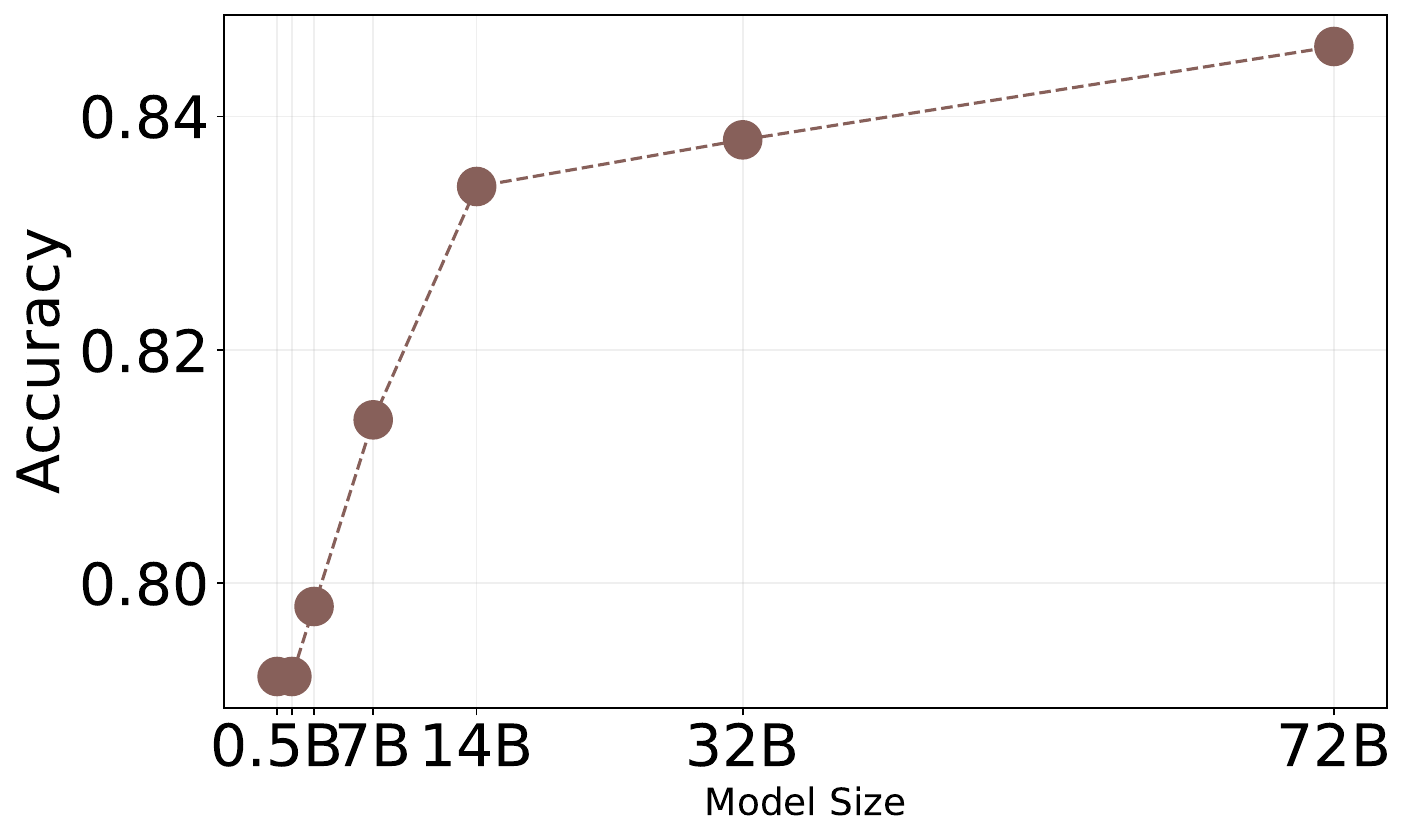}
        \label{fig:training_loss_tokens_lr_v3}
    \end{minipage}  
    % \hfill
    % \begin{minipage}{0.24\linewidth} 
    %     \centering
    %     \includegraphics[width=\linewidth]{figure/N_16.pdf}
    %      \label{fig:training_loss_steps_1B}
    % \end{minipage}  
    % \hfill
    % \begin{minipage}{0.24\linewidth} 
    %     \centering
    %     \includegraphics[width=\linewidth]{figure/2N_16.pdf}
    %      \label{fig:training_loss_steps_1B}
    % \end{minipage}  
    \vspace{-0.25in}
    \caption{Comparison of the performance of PRMs with varying model sizes during inference using the same language model. As model size increases, accuracy improves rapidly, indicating a better capability to capture the complexity necessary for solving mathematical problems.}
    \label{fig:PRM_Models1}
\end{figure*}

% \begin{figure*}[h!]
%     \centering
%     \vspace{-0.15in}
%     \begin{minipage}{0.4\linewidth} 
%         \centering
%         \includegraphics[width=\linewidth]{figure/prm_last_vote_training2.pdf}
%         \label{fig:training_loss_tokens_lr_v3}
%     \end{minipage}  
%     % \hfill
%     \begin{minipage}{0.4\linewidth} 
%         \centering
%         \includegraphics[width=\linewidth]{figure/prm_last_vote_training2-2.pdf}
%          \label{fig:training_loss_steps_1B}
%     \end{minipage}  

%     \vspace{-0.25in}
%     \caption{
%    Impact of PRM Training Steps on Accuracy. 
%    % Figure shows the relationship between training steps and accuracy on Math tasks. 
%    The x-axis represents the number of training steps, while the y-axis indicates the accuracy.}
%     \label{fig:PRM_trainingSteps}
% \end{figure*}

\begin{figure*}[h!]
    \centering
    \vspace{-0.15in}
    \begin{minipage}{0.4\linewidth} 
        \centering
        \includegraphics[width=\linewidth]{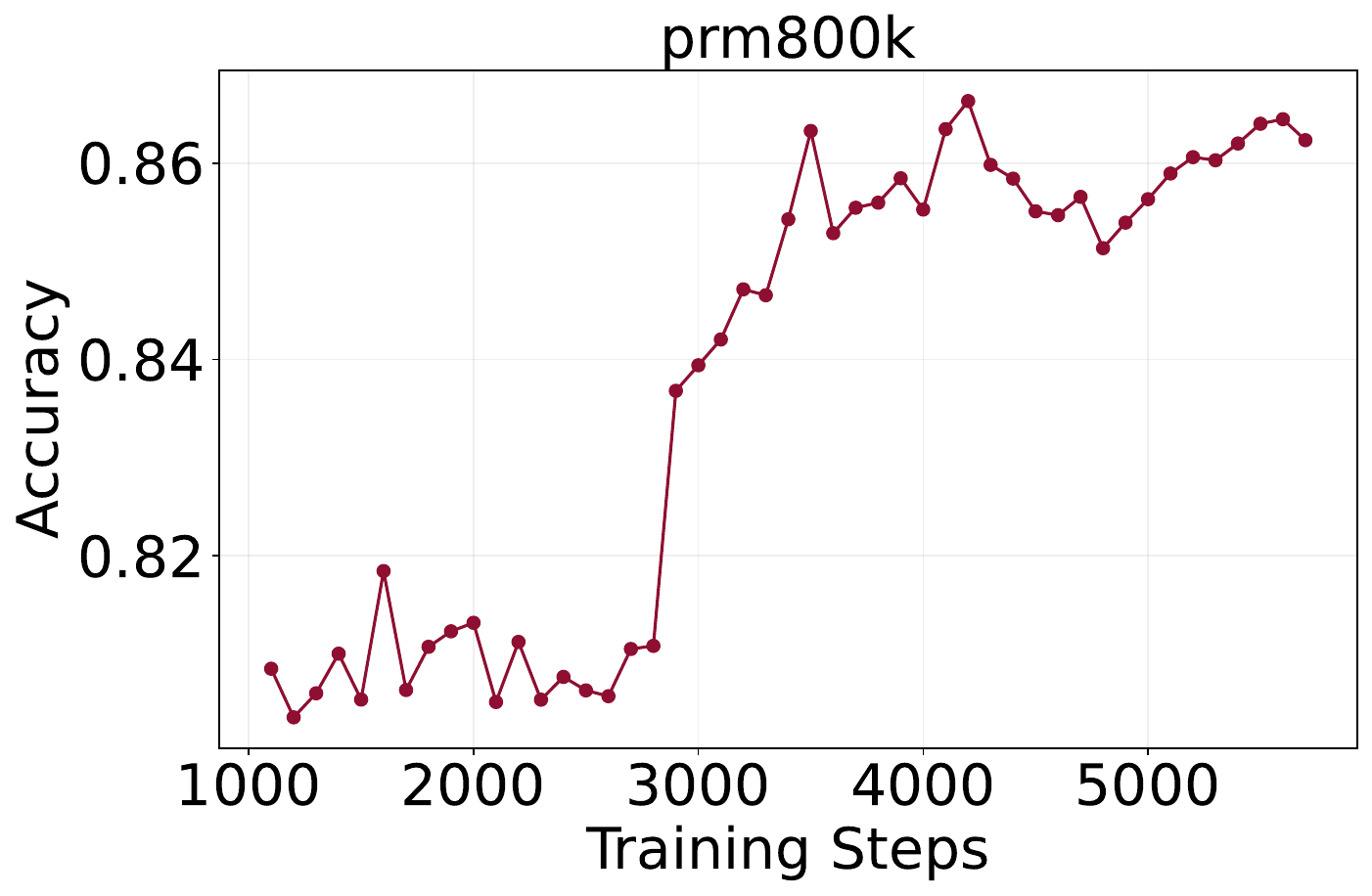}
        \label{fig:training_loss_tokens_lr_v3}
    \end{minipage}  
    % \hfill
    \begin{minipage}{0.4\linewidth} 
        \centering
        \includegraphics[width=\linewidth]{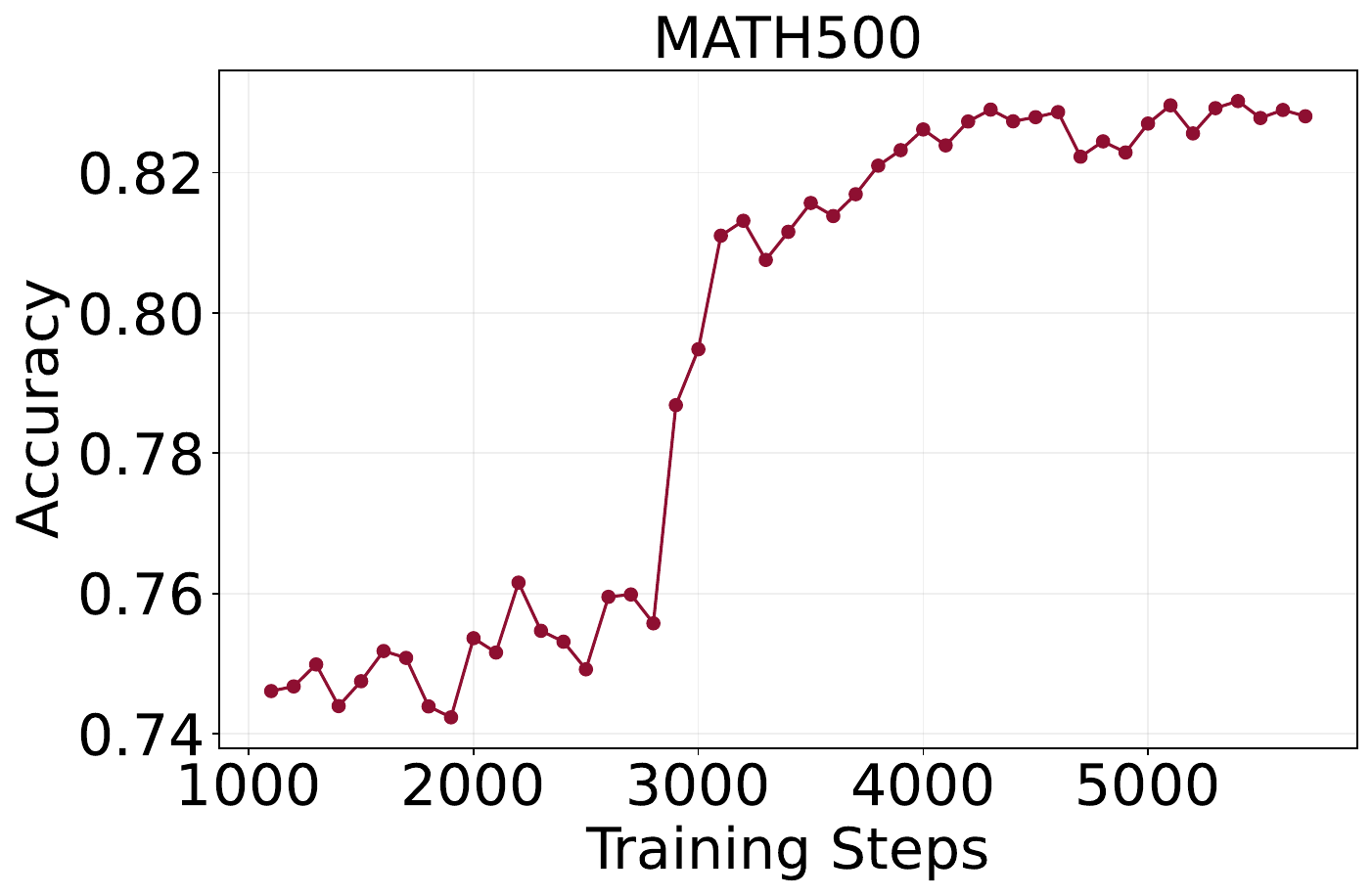}
         \label{fig:training_loss_steps_1B}
    \end{minipage}  

    \vspace{-0.25in}
    \caption{
   Impact of PRM Training Steps on Accuracy. 
   % Figure shows the relationship between training steps and accuracy on Math tasks. 
   The x-axis represents the number of training steps, while the y-axis indicates the accuracy.}
    \label{fig:PRM_trainingSteps}
\end{figure*}

Key findings from our study reveal several insights into the scaling behavior of PRMs. Notably, we observe diminishing returns in model performance as PRM size increases, emphasizing the need for a balanced approach to model size and computational cost. Moreover, the diversity of training datasets significantly influences PRM performance, highlighting the importance of incorporating varied data to enhance accuracy and efficiency.
In addition to training considerations, our research explores test-time scaling strategies \cite{setlur2024rewarding, chen2024simple} to optimize PRM deployment. We evaluate various search strategies, including Best-of-N Sampling, Beam Search, Monte Carlo Tree Search (MCTS), and Majority Voting, to enhance reasoning accuracy in test scenarios. Our results indicate that MCTS is the most effective strategy when ample computational resources are available, while Best-of-N Sampling offers a practical solution under resource constraints due to its simplicity and speed.

Our investigation explores the generalization capabilities of PRMs across domains, particularly from mathematical reasoning to code generation. The results demonstrate that PRMs trained on mathematical corpora perform comparably to those optimized on code-centric datasets, suggesting promising cross-domain adaptability.
Here are the contributions of this paper:

1. \emph{Analysis of Training Compute on PRM Performance}:
We analyze the effects of pretraining and reward training compute on PRM efficacy. By experimenting with models of varying sizes and distinct compute allocations, we identify the balance between model size and compute resources, enhancing generalization across tasks.

2. \emph{Test-Time Scaling Performance With Strategies}:
We assess PRM performance during test-time scaling, focusing on the optimization of performance with search strategies. This involves the importance of selecting search strategies based on computational resources and time constraints.

% We assess PRM performance during test-time scaling, focusing on maintaining high performance as model size and task complexity increase. This involves evaluating the model's adaptability to unseen problem instances and cross-domain performance improvements.

% 3. \emph{PRM Generalization Across Domains}:
% We explore PRM generalization by applying them to both mathematical reasoning and code generation tasks. Utilizing various search methods and diverse training datasets, we assess cross-domain performance, revealing the potential for techniques effective in mathematics to excel in coding challenges. 

3. \wyd{\emph{PRM Generalization Across Domains}: We explore PRM generalization in mathematics and coding tasks. With various search methods and diverse training sets, we assess cross-domain performance, revealing the generalization from math to code. Through quantitative analysis of the patterns inherent in the model's output, our findings indicate that the PRM exhibits a preference for selecting content exhibiting similar underlying reasoning patterns.}

\section{Process Reward Modeling}

Process Reward Models are designed to improve the performance of LLMs by providing granular feedback on intermediate steps during the problem-solving process. Unlike traditional models that only evaluate the final output, PRMs assess each step in a reasoning sequence, identifying and addressing errors as they occur. This step-wise evaluation allows for more accurate and reliable solutions, particularly in complex tasks such as mathematical reasoning and code generation.

% The training of a PRM is a crucial aspect of our approach to improve the performance of LLMs by providing granular feedback on intermediate steps in solution generation. 

\subsection{Automatic Step-Level Annotation and Filtering}

In the training process of Process Reward Models, automatic step-level annotation and filtering are crucial steps to enhance model performance. This process involves the following aspects:

1. Collection of Reasoning Tasks:
To construct a high-quality training dataset, we collect a diverse set of reasoning tasks. These tasks cover a wide range of subjects and difficulty levels, ensuring that the PRM can generalize across different types of reasoning processes. Each task is broken down into a sequence of intermediate steps, with each step representing an action or decision that a model might take when generating a solution.

2. Annotation of Steps with LLM:
Each step in the reasoning process is annotated based on its correctness. To facilitate the annotation process, we utilize pre-trained Large Language Models to generate multiple candidate solutions for each problem. By simulating rollouts from each solution, we identify the intermediate steps where errors occur. Heuristic methods, such as Monte Carlo estimation and binary search, are employed to label each step as correct or incorrect. This automated pipeline generates extensive and diverse supervision data, reducing the dependency on costly human annotations and facilitating scalable training of PRMs.

3. Data Filtering:
To improve the quality of the training data, we introduced an ensemble-based filtering mechanism. This mechanism cross-verifies the correctness of reasoning steps using different LLMs. Only those steps where all LLMs agree on the error location are retained for training. This approach mitigates the noise and inaccuracies inherent in using a single model, thereby enhancing the reliability of the training data.

Through these steps, automatic step-level annotation and filtering significantly enhance the efficiency and accuracy of PRM training, laying a solid foundation for the model's performance in complex reasoning tasks.
We validate the effectiveness of Automatic Step-Level Annotation and Filtering (ASLAF) in section \ref{Generalization Ability}.

\subsection{Objective and Framework}

The primary objective of training the PRM is to transform the evaluation of reasoning processes into a supervised learning task. The PRM is designed to predict the correctness of each step in a sequence of reasoning, thereby providing continuous feedback to guide the LLM
towards generating accurate solutions.
The PRMs are initialized from the Qwen2.5 series models, with the original language modeling head replaced by a scalar-value head designed for evaluating reasoning steps.

In this framework, given a problem \( Q \) and a sequence of reasoning steps \( x_1, x_2, \ldots, x_T \), the PRM outputs a probability \( p_t = \text{PRM}(Q, x_1, x_2, \ldots, x_t) \), which represents the likelihood that the step \( x_t \) is correct. The goal of the training process is to minimize the discrepancy between these predicted probabilities and the true binary labels \( y_t \) assigned to each step, where \( y_t \) indicates whether \( x_t \) is correct or not.

The PRM is trained using supervised learning, with the training data comprising problems paired with annotated reasoning steps, each labeled as correct or incorrect. This transforms the problem into a binary classification task. Specifically, the task is to predict the binary label \( y_t \) for each step:

\begin{equation}
    y_t = \text{PRM}(Q, x_1, x_2, \ldots, x_t)
\end{equation}

To measure the prediction error for each step, the binary cross-entropy loss is utilized, defined as:

\begin{equation}
\mathcal{L} = -\frac{1}{N} \sum_{t=1}^{N} [y_t \log(p_t) + (1-y_t) \log(1-p_t)]
\end{equation}

where \( y_t \) is the true label for the \( t \)-th step, and \( p_t \) is the predicted probability that the step is correct. This loss function helps optimize the PRM to align its predictions with the true correctness of steps.

Once trained, PRMs are integrated with LLMs to provide real-time feedback during the reasoning process.
At each step, the LLM generates potential actions (next reasoning steps), which are evaluated by the PRM. The feedback is used to adjust the policy, guiding the model towards more accurate reasoning pathways.

\section{Experimental Setup}

% In this section, we outline the experimental setup employed to evaluate the performance of Process Reward Models (PRMs) under different data construction methodologies. Our experiments aim to assess the comparative effectiveness of different strategies in enhancing the detection of erroneous reasoning steps and improving downstream task performance.

In this section, we outline the experimental setup employed to evaluate the performance of PRMs under different data construction methodologies. Our experiments aim to assess the comparative effectiveness of different strategies in enhancing the detection of erroneous reasoning steps and improving downstream task performance.

\subsection{Datasets}

\textbf{Post Training} 
We construct the three mathematical training sets and code training sets for PRMs. The mathematical training corpus is derived from PRM 800K~\citep{lightman2023lets} and Math-shepherd~\citep{wang2024math}, which include reasoning processes and evaluation labels. Moreover, we use ASLAF to combine and filter PRM 800K and Math-shepherd datasets. The code corpus is obtained from TACO~\citep{li2023taco}, APPS~\citep{hendrycksapps2021}, and CodeContent~\citep{li2022competition}. We use pre-trained LLM to generate reasoning processes and resulting code, and subsequently employ LLM to annotate them.

We construct three mathematical training sets for PRMs. The mathematical training corpus is derived from PRM 800K (Lightman et al., 2023) and Math-shepherd (Wang et al., 2024), which include reasoning processes and evaluation labels. Moreover, we use our proposed Automatic Step-Level Annotation and Filtering (ASLAF) method to combine and filter samples from PRM800k and Math-shepherd datasets.

The ASLAF dataset contains approximately 1.2 million samples, while PRM800k and Math-Shepherd contain 800k and 445k samples respectively. To ensure fair comparison in our experiments, we randomly sampled 400k examples from each dataset for training. This controlled sampling approach allows us to isolate the effects of data quality and diversity from those of data quantity.

\textbf{Evaluation}
To comprehensively evaluate the model's capability in mathematical reasoning problems, we utilize the MATH-500, American Invitational Mathematics Examination (AIME)~\citep{MAA2024}, and PRM 800K~\cite{lightman2023lets}. In terms of code generation, we employ HUMANEVAL+, MBPP+, LiveCodeBench, and BigCodeBench. EvalPlus~\citep{liu2023your} introduces HUMANEVAL+, which adds 80 specially designed problems to the original HUMANMEVAL~\citep{chen2021evaluating}, along with corrections to the answers in the original dataset. Similarly, MBPP+ provides an additional 35 test cases for MBPP~\citep{austin2021program}.

% apps、taco、code_contests4

% https://huggingface.co/datasets/codeparrot/apps

% https://huggingface.co/datasets/BAAI/TACO

% https://huggingface.co/datasets/deepmind/code_contests

% https://arxiv.org/pdf/2502.01456 APPSMath-Shepherd [Wang et al., 2024a]PRM800k dataset [Lightman et al., 2023],

\subsection{Evaluation Methodology}

To evaluate the efficacy of our trained PRMs, we focus on two primary aspects: 
1) Downstream Task Performance: The models' ability to enhance overall problem-solving effectiveness.
2) Erroneous Step Identification: Precision in detecting specific reasoning errors.

% \textbf{Best-of-N Sampling Strategy}.
Consistent with prior work \citep{lightman2023lets, wang2024math,luo2024improve, cobbe2021gsm8k, yang2024qwen2}, we employ the BON sampling strategy; the highest-scored response from \(N\) candidates is selected according to the PRM, with evaluations specifically using "prm@$N$". 
% We sampled eight responses from Qwen2.5-Math-7B-Instruct and evaluated across various benchmarks like GSM8K, MATH, Minerva Math, GaoKao 2023 En, OlympiadBench, College Math, and MMLU STEM. Each candidate was scored by evaluating the product of individual step scores, as indicated in \citep{lightman2023lets}.

Additionally, the results of majority voting among the eight samples (maj@N) serve as a baseline, while pass@N (the proportion of test instances where any of the eight samples achieved a correct final answer) is referenced as an upper bound.

Through this setup, we aim to glean insights into the comparative strengths and limitations of different annotation methods and their impact on the generalization capabilities of PRMs.

\section{Generalization Ability of Process Reward Model}\label{Generalization Ability}

\begin{figure*}[h!]
    \centering
    \vspace{-0.15in}
    \begin{minipage}{0.4\linewidth} 
        \centering
        \includegraphics[width=\linewidth]{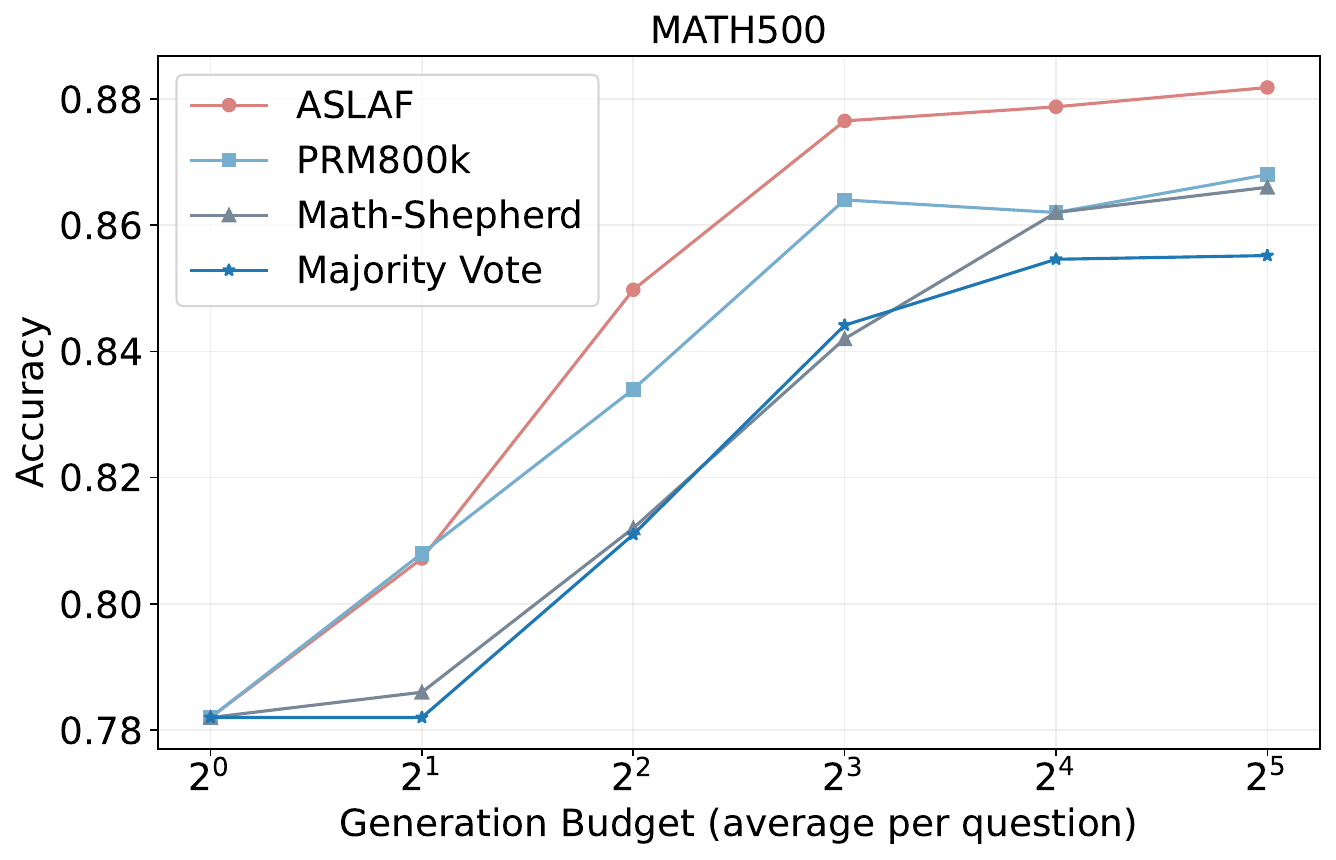}
        \label{fig:training_loss_tokens_lr_v3}
    \end{minipage}  
    % \hfill
    \begin{minipage}{0.4\linewidth} 
        \centering
        \includegraphics[width=\linewidth]{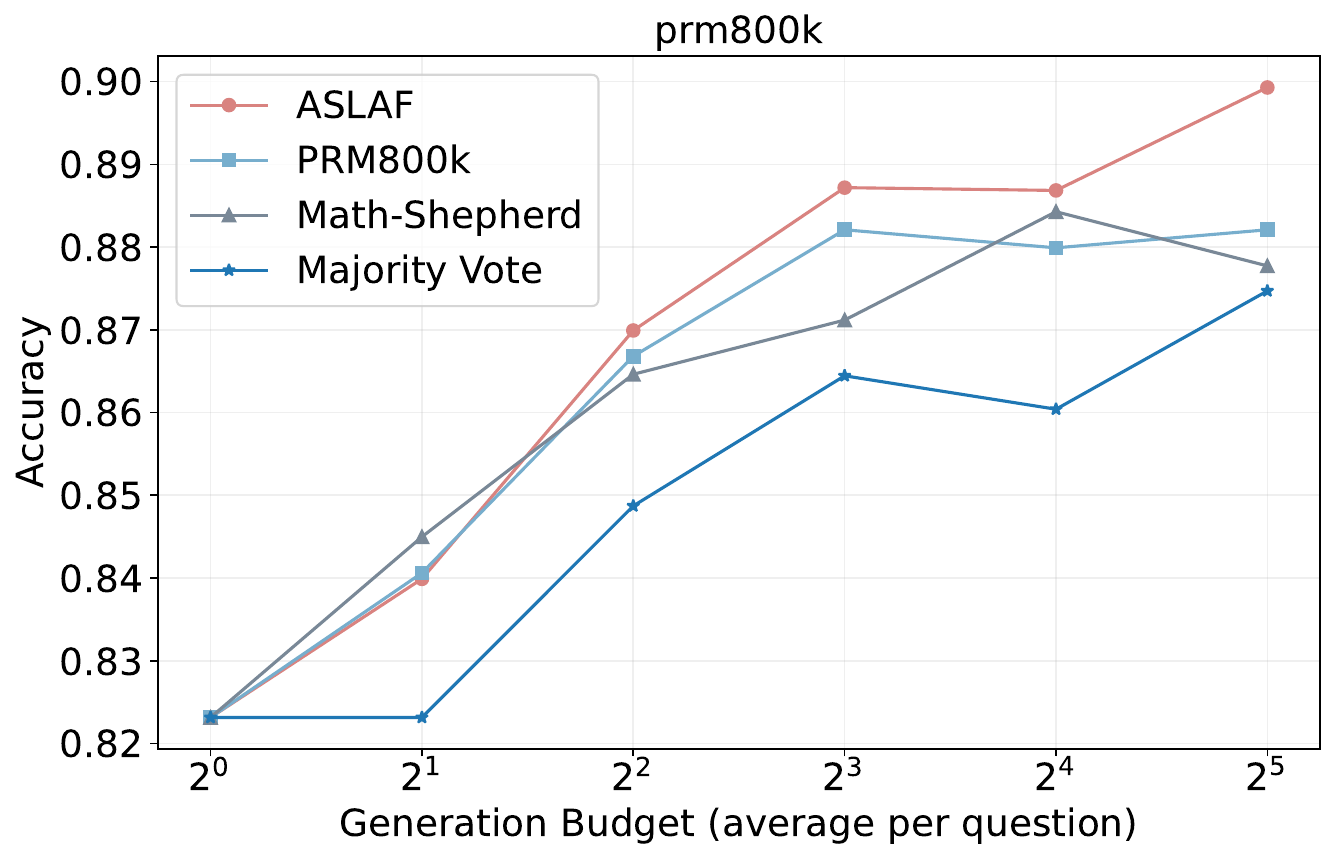}
         \label{fig:training_loss_steps_1B}
    \end{minipage}  

    \vspace{-0.25in}
    \caption{Performance of PRM Training on Different Datasets.}
    \label{fig:PRM_datasets}
\end{figure*}

% \begin{figure}[h!]
%     \centering
%     \vspace{-0.15in}
%     \begin{minipage}{0.8\linewidth} 
%         \centering
%         \includegraphics[width=\linewidth]{figure/prm_min_vote_log2-3.pdf}
%         \label{fig:training_loss_tokens_lr_v3}
%     \end{minipage}  
%     % \hfill
%     \begin{minipage}{\linewidth} 
%         \centering
%         \includegraphics[width=\linewidth]{figure/prm_min_vote_log2-4.pdf}
%          \label{fig:training_loss_steps_1B}
%     \end{minipage}  

%     \vspace{-0.25in}
%     \caption{Performance of PRM Training on Different Datasets.}
%     \label{fig:PRM_datasets}
% \end{figure}

\subsection{Impact of Pre-Training and Reward Training FLOPs}

Understanding the relationship between computational resources and model performance is crucial for optimizing the training of PRMs. We explore the scaling performance for training the PRM, focusing on both pre-training FLOPs, which involve different model sizes, and reward model training FLOPs, which pertain to the specific training needs of the PRM.

The training FLOPs for the PRM consist of two main components:
pre-training FLOPs and reward model training FLOPs.
Pre-training FLOPs involves selecting different model sizes for the PRM, which affects the initial computational requirements. Larger models generally require more FLOPs during the pre-training phase, but they also have the potential to capture more complex patterns in the data.
Reward model training FLOPs per-tains to the specific computational needs during the training of the reward model itself.

\begin{AIbox}{Observation 1}
\label{observe:returns}
\textbf{Diminishing Returns of PRM Scaling.} Larger models initially provide substantial accuracy improvements, effectively capturing the complexity required for solving mathematical problems. However, as the model size continues to increase, the rate of accuracy improvement slows, indicating diminishing returns. This trend suggests that beyond a certain point, further increasing model size results in less significant performance gains relative to the additional computational cost.
\end{AIbox}

Pre-training FLOPs are primarily determined by the size of the model selected for initial training. We use Qwen2.5s to study this problem, including pre-trained models of 7 sizes, including 0.5B, 1.5B, 3B, 7B, 14B, 32B, and 72B.
Figure \ref{fig:PRM_Models1} illustrates the relationship, with the x-axis representing different model sizes of the PRM and the y-axis showing the accuracy on Math tasks, including PRM 800k and MATH-500 datasets. Key observations include:
As the model size increases, accuracy initially improves rapidly, suggesting that larger models better capture the complexity required for solving mathematical problems. 
This quick improvement underscores the benefits of enlarging model size as initialization for PRMs. However, as the model size continues to expand, the rate of accuracy improvement decelerates, reflecting diminishing returns. This pattern implies that beyond a certain threshold, further enlarging the model yields less substantial performance improvements compared to the additional computational expense.
%As the model size increases, accuracy initially improves rapidly, suggesting that larger models better capture the complexity required for solving mathematical problems. This rapid improvement highlights the effectiveness of scaling up model size during pre-training. However, as the model size continues to grow, the rate of accuracy improvement slows, indicating diminishing returns. This trend suggests that beyond a certain point, further increasing model size results in less significant performance gains relative to the additional computational cost.

Reward model training FLOPs refer to the computational resources required during the PRM-specific training phase. Understanding these resources is crucial for optimizing the efficiency and effectiveness of the PRM.
Figure \ref{fig:PRM_trainingSteps} and \ref{fig:PRM_datasets} provide insights into this aspect.

Figure \ref{fig:PRM_trainingSteps} shows the relationship between training steps and accuracy on Math tasks.
As the number of training steps increases, accuracy initially remains unchanged during the early training phase. This plateau is followed by a sudden leap in accuracy, known as the "emergence" phenomenon, after which accuracy continues to increase gradually. This pattern suggests that a critical number of training steps is required before the PRM begins to effectively leverage the information in the data, resulting in a rapid improvement in performance. The subsequent gradual increase indicates that further training continues to refine the PRM's capabilities, albeit at a slower pace.

\begin{AIbox}{Observation 2}
\label{observe:pareto}
\textbf{The choice and diversity of training datasets significantly impact the performance of Process Reward Models. }
It highlights the importance of incorporating a wide range of data during reward model training to maximize accuracy and efficiency, ultimately leading to improved reasoning performance of LLMs with optimized computational resources.
\end{AIbox}

Figure \ref{fig:PRM_datasets} shows the impact of different PRM training on various datasets during inference. The x-axis represents the BON strategy with varying N, and the y-axis indicates the accuracy on Math tasks. The training datasets include the PRM800k dataset, the math-shepherd dataset, and the ASLAF dataset based on both. 
It highlights that PRM trained on our proposed ASLAF outperforms PRMs trained on the PRM800k and Math-Shepherd datasets, as well as the majority vote method.

The key observations are: \emph{Dataset Influence.} The choice of training dataset significantly affects model performance. The ASLAF dataset generally provides higher accuracy, suggesting that high-quality and diverse datasets can enhance training efficiency and effectiveness.
  \emph{BON Strategy Impact.} As N increases in the BON strategy, accuracy improves, demonstrating the utility of evaluating multiple candidate solutions to identify the most accurate reasoning path.
  \emph{Trade-offs in Training and Testing FLOPs.} While increasing N in the BON strategy can enhance accuracy, it also requires more computational resources, underscoring the need for a balanced approach to optimize training FLOPs.

The figure reveals that the choice of training dataset significantly impacts model performance. Using the ASLAF method to annotate step labels of PRM800k and math-shepherd results in the highest accuracy across different N values, indicating that a diverse and comprehensive training dataset enhances the model's generalization ability and performance on complex tasks. This underscores the importance of incorporating a wide range of data during reward model training to maximize accuracy.

% \textbf{Lack of exp:}
% Key observations include:

% 1. Diminishing Returns: Initial increases in training FLOPs result in substantial performance gains, but the benefits diminish beyond a certain threshold. This indicates an optimal range of FLOPs for training PRMs, beyond which additional resources yield marginal improvements.

% 2. Impact of Training Techniques: Techniques such as consensus filtering and ensemble methods significantly affect the training FLOPs scaling behavior. These methods enable models to achieve high performance with reduced computational resources, optimizing the trade-off between FLOPs and accuracy.

% 3. Data Efficiency: High-quality, diverse datasets enhance training efficiency, allowing PRMs to achieve better performance with fewer FLOPs. This underscores the importance of using consensus-filtered datasets to maximize training efficiency.

\begin{figure*}[h!]
    \centering
    \vspace{-0.15in}
    \begin{minipage}{0.4\linewidth} 
        \centering
        \includegraphics[width=\linewidth]{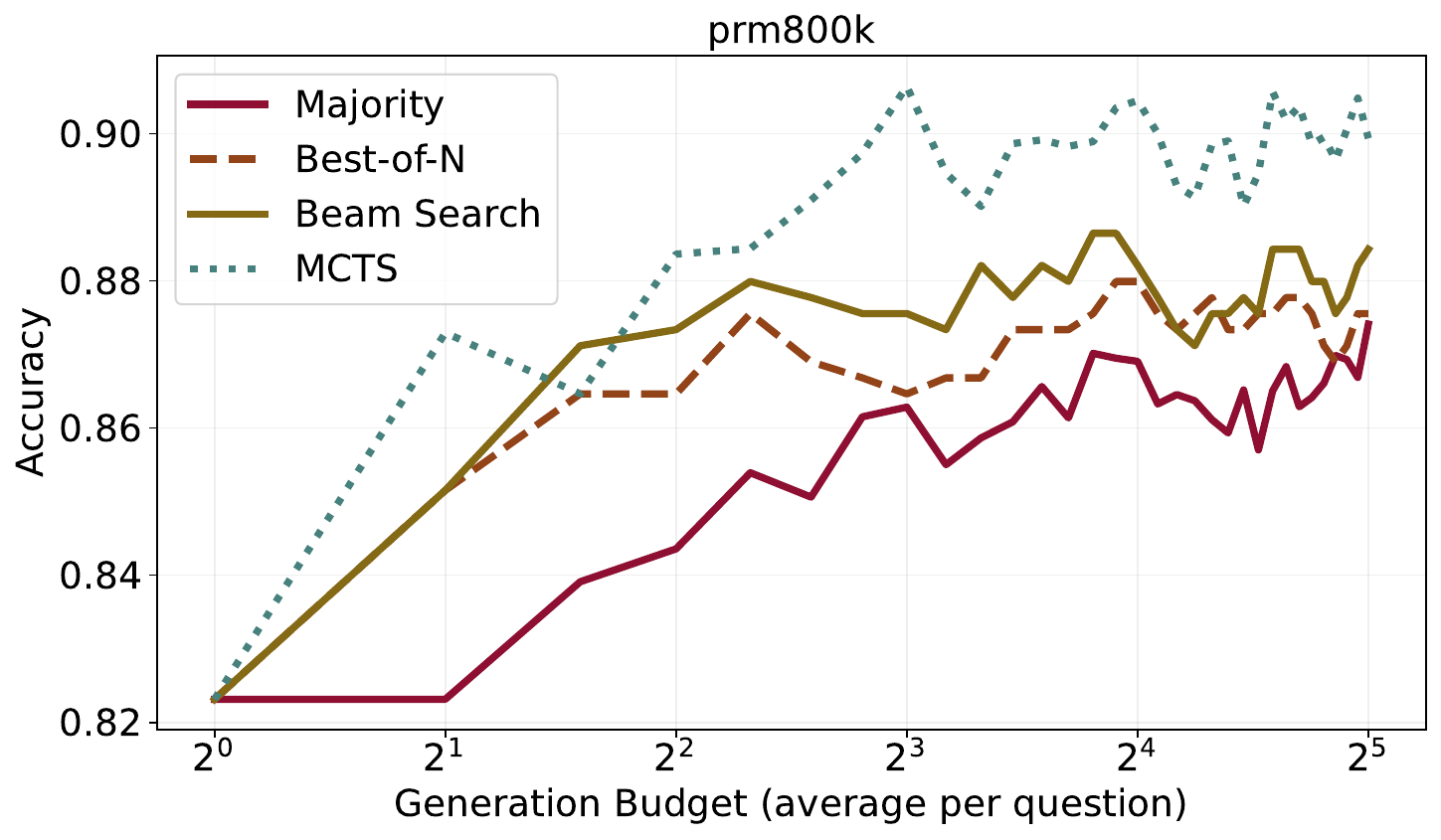}
        \label{fig:training_loss_tokens_lr_v3}
    \end{minipage}  
    % \hfill
    \begin{minipage}{0.4\linewidth} 
        \centering
        \includegraphics[width=\linewidth]{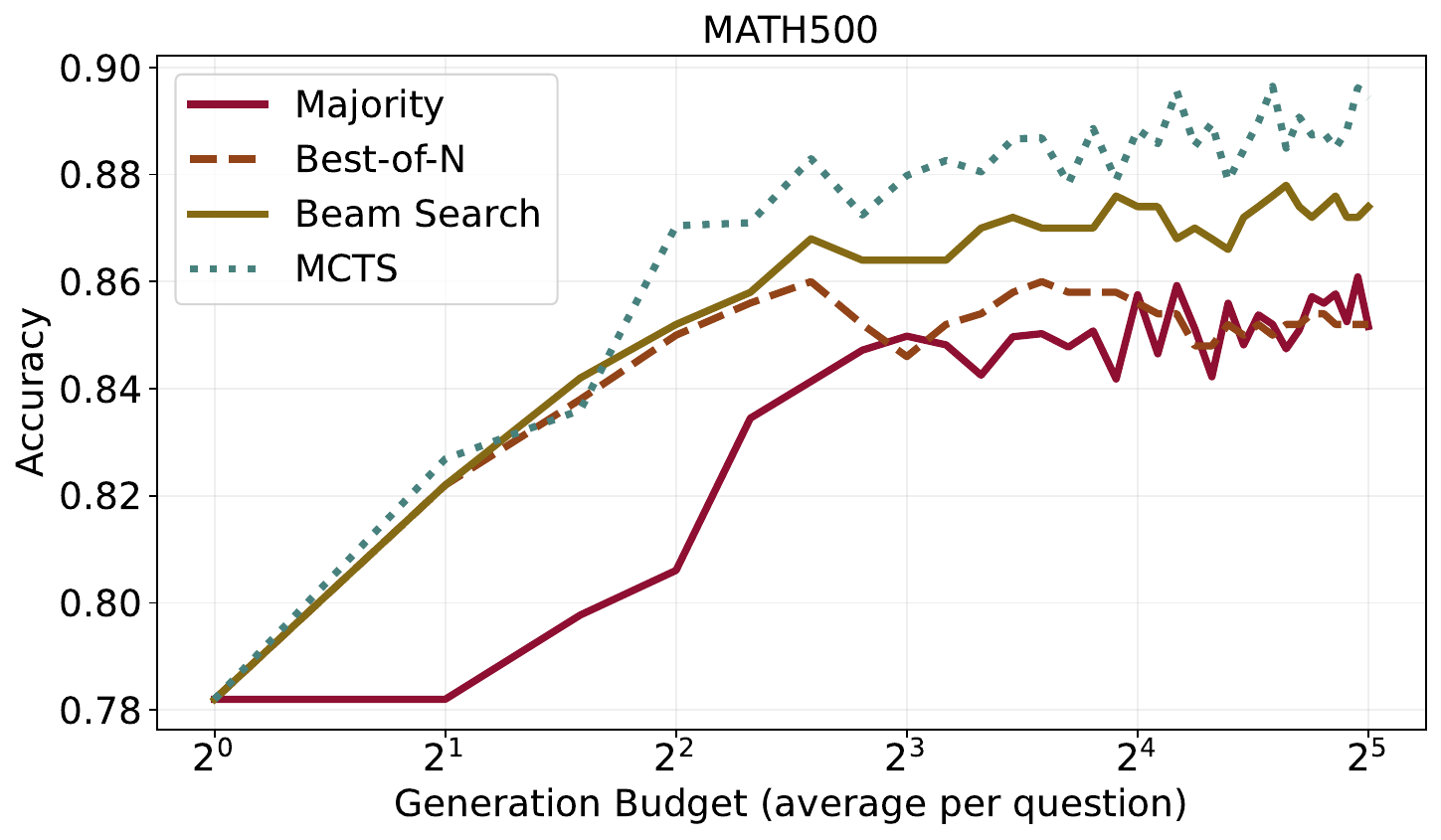}
         \label{fig:training_loss_steps_1B}
    \end{minipage}  

    \vspace{-0.25in}
    \caption{ Impact of Different Search Strategies on Accuracy with Generation Budget. This figure examines the effect of various search strategies on accuracy, with the x-axis representing the generation budget (average per question) and the y-axis showing accuracy on Math tasks, including PRM800k and Math500. The search strategies analyzed include best-of-N, beam search, MCTS, and majority vote.}
    \label{fig:Generation}
\end{figure*}

\begin{figure*}[h!]
    \centering
    \vspace{-0.15in}
    \begin{minipage}{0.4\linewidth} 
        \centering
        \includegraphics[width=\linewidth]{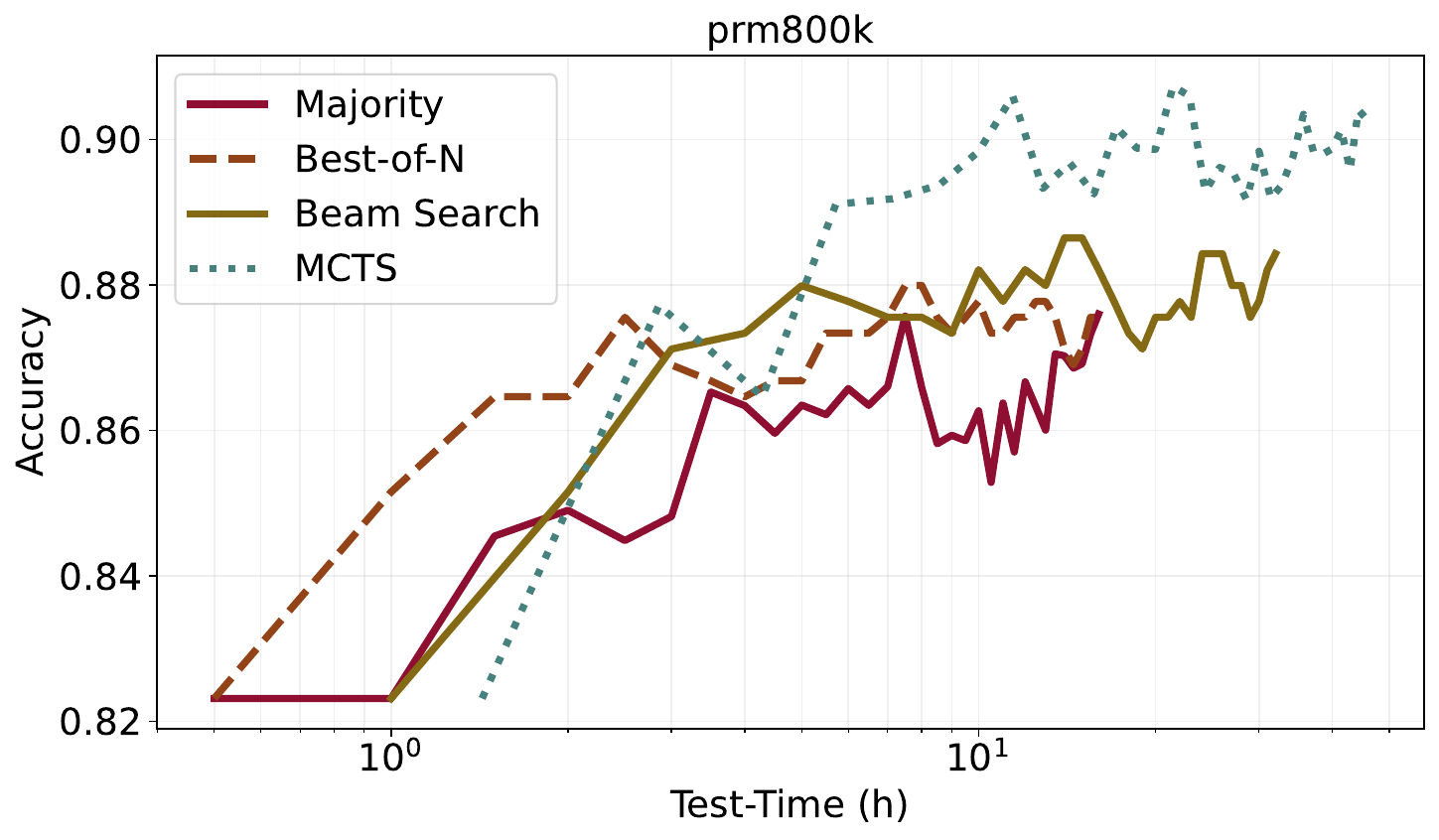}
        \label{fig:training_loss_tokens_lr_v3}
    \end{minipage}  
    % \hfill
    \begin{minipage}{0.4\linewidth} 
        \centering
        \includegraphics[width=\linewidth]{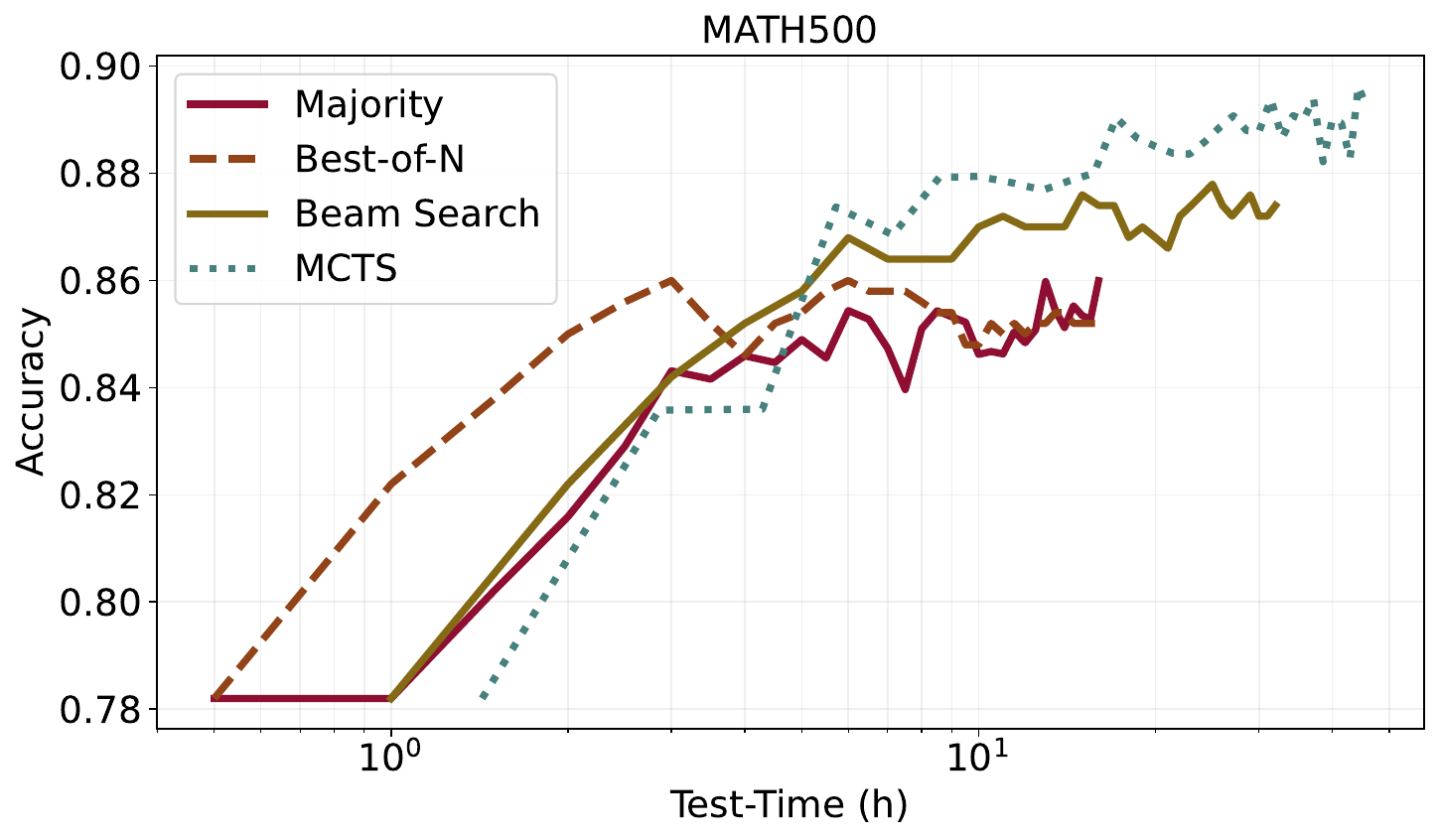}
         \label{fig:training_loss_steps_1B}
    \end{minipage}  

    \vspace{-0.25in}
    \caption{ Impact of Different Search Strategies on Accuracy with Test-Time Constraints. This figure explores the performance of different search strategies under fixed test-time constraints, with the x-axis representing Test-Time (hours) and the y-axis showing accuracy on Math tasks, including PRM800k and Math500. The strategies include best-of-N, beam search, MCTS, and majority vote.}
    \label{fig:testtime}
\end{figure*}

\subsection{Test-Time Scaling Performance}
Test-time scaling involves optimizing the deployment of PRMs to maximize their performance across various test scenarios. This includes strategies for efficient candidate generation, scoring, and selection, as well as methods for leveraging multiple PRMs to achieve consensus and improve accuracy.
To enhance the reasoning process at test time, we employ search strategies that utilize the PRM's feedback to navigate the solution space effectively. 
The key strategies include:

\textbf{Best-of-N Sampling}: Generate \( N \) candidates and select the highest-scoring solution according to a preference reward model:
$\text{Best Solution} = \arg \max_{i \in \{1, 2, \ldots, N\}} R_i$
This approach improves solution quality by exploring diverse reasoning paths.

% Generate \( N \) candidate solutions for a given problem. Each candidate solution is evaluated using the PRM, which assigns a reward \( R_i \) to each solution \( i \). The solution with the highest reward is selected:
% $ \text{Best Solution} = \arg \max_{i \in \{1, 2, \ldots, N\}} R_i $
% This method increases the likelihood of finding a correct solution by exploring diverse reasoning paths.

\textbf{Beam Search}: 
Maintain \( K \) highest-scoring partial solutions at each step. For each path, calculate cumulative scores:
$\text{Top Paths} = \text{Top-K}\left\{ \sum_{t=1}^{T} R(x_t) \right\}$
This efficiently balances exploration and computational resource allocation.

% Start with an initial set of paths (beam width \( K \)). At each step, expand each path by generating possible next steps. Use the PRM to score each expanded path. Retain the top \( K \) paths with the highest cumulative scores: $ \text{Top Paths} = \text{Select Top K from} \left\{ \sum_{t=1}^{T} R(x_t) \right\} $
% This approach balances exploration and exploitation, focusing computational resources on the most promising paths.

\textbf{Monte Carlo Tree Search (MCTS)}: Represent the reasoning process as a tree where nodes are states and edges are actions.
Use a policy, such as Upper Confidence Bound for Trees (UCT), to select the next node:
% \begin{equation}
$    \text{UCT}(s, a) = Q(s, a) + c \cdot \sqrt{\frac{\ln N(s)}{N(s, a)}}$
% \end{equation}
where \( Q(s, a) \) is the estimated value of action \( a \) from state \( s \), \( N(s) \) is the visit count of state \( s \), and \( N(s, a) \) is the visit count of action \( a \) from state \( s \).
Perform rollouts to simulate the outcome of following a particular path.
Update the value estimates of nodes based on the results of the rollouts.
MCTS effectively balances exploration and exploitation, allowing the model to explore vast solution spaces strategically.

\textbf{Majority Voting}: Generate multiple candidate solutions. Aggregate the final answers, selecting the most frequently occurring answer:
$ \text{Final Answer} = \text{Mode}(\{A_1, A_2, \ldots, A_N\}) $
Majority voting leverages the collective insights from multiple solutions, enhancing the robustness of the final answer.

% We evaluated the effectiveness of our test-time scaling strategies using a suite of mathematical benchmarks, including GSM8K, MATH, Minerva Math, GaoKao 2023 En, OlympiadBench, College Math, and MMLU STEM. %The evaluation metrics included the Best-of-8 accuracy (prm@8), majority voting accuracy (maj@8), and the proportion of test samples where any of the eight samplings led to the correct final answers (pass@8).

We evaluate the effectiveness of test-time scaling strategies using a suite of mathematical benchmarks. These experiments demonstrate significant improvements in reasoning accuracy, underscoring the practical utility of test-time scaling strategies.

\begin{table*}[h!]
\centering
\begin{tabular}{lc|cccc}

\toprule
Reward Model & N & HumanEval+ & MBPP+ & LiveCodeBench & BigCodeBench \\
\midrule
\multicolumn{4}{l}{\textit{Generate Model: Qwen2.5-Coder-7B-Instruct}}\\
\midrule
- & 1 & 81.1 & 72.2 & 14.2 & 49.0  \\
Majority Vote & 8 & 81.5 & 73.7 & 15.6 & 50.7 \\
% Qwen2.5-72B-PRM & 8 & 82.9 & 75.1 & 18.7 & 55.1 \\
PRM-Math & 8 & 86.0 & 77.3 & 22.6 & 54.2\\
PRM-Code & 8 & 84.2 & 78.6 & 18.7 & 52.3\\
\midrule
\multicolumn{4}{l}{\textit{Generate Model: Qwen2.5-Coder-32B-Instruct}}\\
\midrule
- & 1 & 82.9 & 77.0 & 25.8 & 57.5  \\
Majority Vote & 8 & 85.8 & 77.6 & 28.2 & 59.4 \\
PRM-Math & 8 & 91.5 & 78.8 & 30.3 & 60.1\\
PRM-Code & 8 & 89.0 & 78.4 & 29.7 & 60.1\\
\midrule
\multicolumn{4}{l}{\textit{Generate Model: QwQ-32B-Preview}}\\
\midrule
- & 1 & 84.8 & 70.9 & 36.1 & 54.3  \\
Majority Vote & 4  & 86.8 & 76.0 & 40.1 & 54.6 \\
PRM-Math & 4 & 89.6 & 76.2 & 41.9 & 54.2\\
PRM-Code & 4 & 86.0 & 79.6 & 41.9 & 53.1 \\
% Qwen2.5-72B-PRM & 8 & 85.4 & 78.6 & 47.6 & 54.2\\
% Qwen2.5-72B-mergedata & 8 & 82.3 & 76.2 & 47.6 & 54.2\\

% \midrule
 
% Qwen2.5-Coder-32B-Instruct & 200 & 0.80 & 0.82 & 0.83 & 0.81 \\
% QwQ-32B-Preview & 150 & 0.90 & 0.91 & 0.89 & 0.92  \\
\bottomrule
\end{tabular}
\caption{The performance of PRM models trained on different datasets for code generation tasks varies across different generative models. All the models are trained from Qwen2.5-72B. Campared to PRM trained with Math, the PRM model trained on code datasets does not exhibit a significant advantage in the task. 
% This observation suggests that the generalization capability of reasoning skills derived from mathematical datasets is more robust.
}
\label{tab:model_results}
\end{table*}

\begin{AIbox}{Observation 3}
\label{observe:pareto}
\textbf{Select search strategies to optimize the performance of Process Reward Models based on the specific computational context.} MCTS is identified as the most effective search strategy when there is ample computational resource availability, outperforming beam search, best-of-N, and majority vote. Best-of-N initially outperforms other methods under limited test-time conditions due to its simplicity and speed.
\end{AIbox}

Figure \ref{fig:Generation} examines the effect of various search strategies on accuracy with the generation budget. 
As the generation budget increases, MCTS consistently outperforms other strategies, followed by beam search, best-of-N, and majority vote in descending order of accuracy. This indicates that MCTS is the most effective strategy when ample computational resources are available, as it balances exploration and exploitation efficiently. Beam search also performs well, offering a good trade-off between resource usage and accuracy. Best-of-N and majority vote, while simpler, are less effective in maximizing accuracy under higher generation budgets.
Majority Voting enhances robustness, particularly in scenarios with high variability in candidate solutions.
Search strategies with PRM such as MCTS, beam search, best-of-N outperform majority vote, achieving higher accuracy and better error localization capabilities. Integrating the PRM into the decoding process leads to substantial improvements in reasoning accuracy, as the model receives continuous feedback and refines its reasoning iteratively.
These findings underscore the importance of test-time scaling in enhancing the practical utility of PRMs.

Moreover, Figure \ref{fig:testtime} explores performance under fixed test-time constraints. Unlike the generation budget scenario, best-of-N initially outperforms other methods under limited test-time conditions due to its simplicity and speed. However, as more test-time is allocated, MCTS surpasses other strategies, demonstrating its superior ability to explore the solution space effectively given sufficient time. Beam search also shows competitive performance but requires more time to achieve results comparable to MCTS. Majority vote remains the least effective strategy under time constraints, highlighting its limitations in leveraging extensive search.

These findings underscores the importance of selecting search strategies based on available computational resources and time constraints.

\subsection{Generalization Across Domains}

% To study the generalization ability of PRMs, we train PRMs in both math and code corpus respectively and evaluate them in code generation tasks.2
To investigate the generalization capabilities of PRMs, we independently train PRMs on domain-specific corpora (math and code datasets) and rigorously assess their performance on cross-domain code generation benchmarks.

% \textbf{PRM for Code}
% We hypothesize that the methodologies effective in mathematical reasoning can be adapted to the domain of code generation. 

% % Formally, let \( \theta \) represent the PRM parameters fine-tuned on mathematical reasoning tasks. We define a fine-tuned model \( \theta^{\text{code}} \) as follows:
% Formally, let \( \theta \) represent the PRM parameters. We define a fine-tuned model \( \theta^{\text{code}} \) as follows:

% $$
% \theta^{\text{code}} = \arg\min_\theta \sum_{i=1}^N \mathcal{L}_{\text{code}}(\theta, x_i^{\text{code}}, y_i^{\text{code}})
% $$

% where \( \mathcal{L}_{\text{code}} \) is the loss function for code generation tasks, \( x_i^{\text{code}} \) are the input coding problems, and \( y_i^{\text{code}} \) are the corresponding correct solutions.

\begin{AIbox}{Observation 4}
\textbf{Generalization from Math to Code.}
PRMs trained on math corpora demonstrate comparable performance to those optimized on code-centric datasets in code generation tasks.

% PRMs post-trained with Math corpus are competitive with those on Code corpus in code generative tasks.

\end{AIbox}

% As the results shown in Table~\ref{tab:model_results}, PRMs consist promote the model performance in most tasks. Specificly, LLMs with PRM gain 4\% upgrade in most tasks. For the PRM trained with different corpus, Merge Data present best performance, in constract to the results in Math reasoning tasks. Compared with PRM trained with code-centric datasets, Merge data perform even better, which probabily casued by the high quaility label from PRM and Math-shepherd. 

\begin{figure}[t]
    \centering
    \small
    % \resizebox{0.8\textwidth}{!}{
\begin{multiqa}
    \begin{questionlayer}
You are given a string s and an integer k. Encrypt the string using the following algorithm:
For each character c in s, replace c with the $k^{th}$ character after c in the string (in a cyclic manner).
Return the encrypted string.

...

% Constraints:

% $1 <= s.length <= 100$, 
% $1 <= k <= 10^4$, 
% s consists only of lowercase English letters.
    \end{questionlayer}
%     \begin{questionlayer}
    \begin{reasonlayer}
\textbf{Step 1}:\begin{verbatim}...
# Find the k-th character after 'char' 
    in a cyclic manner 
new_char = s[(s.index(char) + k) % 
    length]
...\end{verbatim}
\begin{flushright}
\textcolor{red}{$0.5371$}
\end{flushright}
% \textbf{Step 1}: \begin{verbatim}class Solution:
%     def getEncryptedString(self, s: str, k: int) -> str:
%         encrypted_string = ""
%         length = len(s)
%         for char in s:
%             # Find the k-th character after 'char' in a cyclic manner 
%             new_char = s[(s.index(char) + k) % length]
%             encrypted_string += new_char
%         return encrypted_string\end{verbatim}
% \begin{flushright}
% \textcolor{red}{$0.5371$}
% \end{flushright}
\textbf{Step 2}:

However, the above solution uses `s.index(char)' which will not work correctly if there are duplicate characters in the string `s'. Instead, we use current index of the character in the loop. Here is the corrected solution:
\begin{flushright}
\textcolor{red}{$0.7563$}
\end{flushright}
\textbf{Step 3}:\begin{verbatim}...
# Find the k-th character after the 
current character in a cyclic manner
new_char = s[(i + k) % length]
...\end{verbatim}
\begin{flushright}
\textcolor{red}{$0.875$}
\end{flushright}
% \textbf{Step 3}: \begin{verbatim}class Solution:
%     def getEncryptedString(self, s: str, k: int) -> str:
%         encrypted_string = ""
%         length = len(s)
%         for i in range(length):
%             # Find the k-th character after the current character in a cyclic manner
%             new_char = s[(i + k) % length]
%             encrypted_string += new_char
%         return encrypted_string\end{verbatim}
% \begin{flushright}
%         \textcolor{red}{$0.875$}
% \end{flushright}
    % \textbf{Step 1}: To find the remainder of 2004 divided by 12, we can use the division algorithm, which states that for any integers \(a\) and \(b\) (with \(b > 0\)), there exist unique integers \(q\) (the quotient) and \(r\) (the remainder) such that \(a = bq + r\) and \(0 \leq r < b\). \textcolor{red}{$[0.958, 0.988]$} \\

    % \textbf{Step 2}: Here, \(a = 2004\) and \(b = 12\). We need to find \(q\) and \(r\) such that:
    % \[
    % 2004 = 12q + r
    % \]
    % where \(0 \leq r < 12\). \textcolor{red}{$[0.924, 0.987]$} \\

    % \textbf{Step 3}: First, we perform the division of 2004 by 12:
    % \[
    % 2004 \div 12 = 167
    % \]
    % This means the quotient \(q\) is 167.
    %  Now, we can find the remainder \(r\) by multiplying the quotient by the divisor and subtracting the result from the original number:
    % \[
    % 2004 - 12 \times 167 = 2004 - 2004 = 0
    % \]
    % \textcolor{red}{$[0.777, 0.719]$} \\

    % \textbf{Step 4}: Thus, the remainder \(r\) is:
    % \[
    % r = 0
    % \]
    % \textcolor{red}{$[0.777, 0.378]$} \\
    
    % \textbf{Step 5}: Therefore, the remainder when 2004 is divided by 12 is:
    % $\boxed{0}$. \textcolor{red}{$[0.622, 0.349]$}

    \end{reasonlayer}
    % \begin{answerlayer}
    % An integer is divisible by 12 if and only if it is divisible by both 3 and 4. 

    % Because \( 2 + 0 + 0 + 4 = 6 \) is divisible by 3, 2004 is divisible by 3. 
    
    % Also, the last two digits of 2004 form a multiple of 4, so 2004 is divisible by 4 as well. 
    
    % Therefore, 2004 is divisible by 12 and hence leaves a remainder of \( \boxed{0} \) when divided by 12.
    
    % % The remainder is $\boxed{0}$.
    % \end{answerlayer}

\end{multiqa}
% }
    \caption{PRM score for a case with a correct answer. The full context is provided in Figure~\ref{fig-ex2} in the Appendix.}\label{fig-ex1}
\end{figure}

% You are given a string s and an integer k. Encrypt the string using the following algorithm:
% For each character c in s, replace c with the $k^{th}$ character after c in the string (in a cyclic manner).
% Return the encrypted string.\\

% Example 1:\\
% Input: s = "dart", k = 3\\
% Output: "tdar"\\

% Explanation:

% For $i = 0$, the $3^{rd}$ character after 'd' is 't'.
% For $i = 1$, the $3^{rd}$ character after 'a' is 'd'.
% For $i = 2$, the $3^{rd}$ character after 'r' is 'a'.
% For $i = 3$, the $3^{rd}$ character after 't' is 'r'.\\

% Example 2:

% Input: s = "aaa", k = 1\\
% Output: "aaa"\\

% Explanation:\\
% As all the characters are the same, the encrypted string will also be the same.\\

% Constraints:

% $1 <= s.length <= 100$, 
% $1 <= k <= 10^4$, 
% s consists only of lowercase English letters.
%     \end{questionlayer}

\textbf{Generalization from Math to Code Domain}
As evidenced by the experimental results in Table~\ref{tab:model_results}, PRMs consistently enhance model performance across most evaluation tasks. Specifically, LLMs augmented with PRMs trained with math domain, achieve a 4\% average performance improvement in code generation benchmarks. Notably, PRMs trained on ASLAF datasets exhibit superior generalization capabilities compared to those trained solely on code-centric datasets. 

\textbf{\wyd{Pattern Similarity}} 
\wyd{To assess whether PRM captures general reasoning patterns across mathematical and coding tasks, we employ a gradient-based similarity metric to evaluate the internal patterns in responses \citep{wang2025exploring}. 
Specifically, for a reference model \( \mathscr{F} \) and an input sentence \(s\), we define the activation of the $i$-th weight \(\theta^{r}_i\) as:
}

\begin{equation*}
    A(\theta^{r},s)_i = |\theta^{r}_i\cdot \frac{\partial\mathscr{F}(\theta^{r}_i, s)}{\partial \theta^{r}_i}|.
\end{equation*}

\wyd{By concatenating all \(A_i\), we obtain a vector \(A\) that represents the activation patterns of the reference model for the given input. Consequently, we can evaluate the similarity between two reasoning outputs sets \(S_1, S_2\), based on their pattern similarity using the following formula:}

\begin{align*}
\mathscr{S} = \sum_{s_1\in S_1,s_2 \in S_2}\frac{\nabla A(\theta^{r}, s_1)\cdot\nabla A(\theta^{r}, s_2)}{||A(\theta^{r},s_1)||_2,||A(\theta^{r},s_2)||_2}
\end{align*}

\wyd{As the result in Table~\ref{tab:similarity}, the pattern similarity between Math$_{PRM}$ and Code$_{PRM}$ exceeds that of other pairs, suggesting that the PRM model identifies cross-domain responses with similar patterns. For instance, in Figure~\ref{fig-ex1}, responses with rethinking patterns receive higher scores from PRM.  }

% \begin{figure}[t!]
%     \centering
%         \includegraphics[width=0.8\linewidth]{figure/04_math_prm_code.pdf}
%          \label{fig:training_loss_steps_1B}

%     \caption{The performance of Qwen2.5-Coder-32B-Instruct on LiveCodeBench. With PRM trained with math corpus, generate model shows significant advantage compared to the single pass.}
%     \label{fig:math_cate}
% \end{figure}

\begin{table}[t!]
\centering
\begin{tabular}{ll|c}

\toprule
 $S_1$       & $S_2$        & $\mathscr{S}$\\
\midrule
Math$_{PRM}$ & Code$_{PRM}$ &  30.95 \\
Math$_{PRM}$ & Code         &  29.07 \\
        Math & Code$_{PRM}$ &  29.42 \\
        Math & Code         &  26.75 \\
\bottomrule
\end{tabular}
\caption{
% Similarity of response of Qwen2.5-Coder-32B-Instruct across code and math domains. Math refer to the response for math tasks, while Math$_{PRM}$ refer to the responses selected by PRM model. Code and Code$_{PRM}$ is the same meaning.
% Similarity of Qwen2.5-Coder-32B-Instruct responses across code and math domains. Math$_{PRM}$ denotes the responses for math tasks selected by the PRM model, while Math refer to all the responses for math tasks. 
Similarity of responses from Qwen2.5-Coder-32B-Instruct across code and math domains. Here, "Math" denotes responses to math tasks, while "Math$_{PRM}$" are those selected by the PRM-Math; "Code" and "Code$_{PRM}$" follow the same convention.
}
\label{tab:similarity}
\end{table}

% \textbf{Case Study}
% To systematically evaluate cross-domain generalization capabilities, we categorize the problems in LiveCodeBench into four distinct classes: 1) Mathematics, 2) Graph Theory, 3) Data Structure, and 4) Dynamic Programming. The categorization process is conducted using the Qwen2.5-72B model, introduced in Appdenix~\ref{app:Cate_live}. Quantitative analysis, as illustrated in Figure~\ref{fig:math_cate}, reveals that performance improvements are predominantly concentrated in the Mathematics category, with only marginal gains observed in the other domains. %This disparity suggests a potential bias or specialization in the model's ability to generalize across diverse problem types, highlighting the need for further investigation into domain-specific adaptability.

%Furthermore, our investigation uncovers an implementation-level correlation: code solutions employing dual-template architectures (Figure~\ref{fig-ex1}) demonstrate statistically significant score advantages in final evaluation phases. This pattern suggests that hybrid code generation strategies leveraging complementary template structures may enhance robustness during program synthesis.

% \input{05_related_work}

\section{Conclusion}
This paper demonstrates the potential of Process Reward Models to significantly enhance the reasoning capabilities of Large Language Models across diverse domains, particularly in transitioning from mathematical reasoning to code generation tasks. 
Our comprehensive analysis provides key insights into the scalability, efficiency, and adaptability of PRMs, contributing to the broader understanding of their application and development.
Our findings indicate that as PRM size increases, performance improvements follow a pattern of diminishing returns, emphasizing the importance of balancing model size with computational cost. Additionally, the diversity of training datasets significantly impacts PRM performance, highlighting the importance of incorporating varied data in reward model training to improve accuracy and efficiency.

\section{Acknowledge}
This paper is supported by NSFC project 62476009.

\section{Limitations}

In this study, we acknowledge several limitations that may impact the interpretation and generalization of our findings:

1. Computational Resource Constraints: While we explored various scaling strategies and model sizes, our experiments were limited by available computational resources. This constraint may have influenced the extent to which we could explore larger model architectures and more complex scaling strategies.

2. Dataset Diversity: Although we emphasized the importance of diverse training datasets, the scope of our dataset collection was limited to available resources. This may have restricted the breadth of data diversity and, consequently, the generalizability of our findings across all possible domains and tasks.

3. Focus on Mathematical and Code Generation Tasks: Our study primarily focused on mathematical reasoning and code generation tasks. While these domains provide valuable insights into PRM capabilities, further research is needed to explore PRM performance in other domains to fully understand their cross-domain adaptability.

4.  Evaluation Metrics: The metrics used to assess PRM performance were primarily focused on accuracy and efficiency. Future work could benefit from incorporating additional metrics, such as interpretability and robustness, to provide a more comprehensive evaluation of PRM capabilities.

Addressing these limitations in future research will be crucial for further advancing the understanding and development of Process Reward Models, ultimately leading to more robust and versatile applications across various fields.

% Scaling Laws for Reward Model Overoptimization, 

% Scaling Laws for a Multi-Agent Reinforcement Learning Model

% Scaling laws for single-agent reinforcement learning, 

% scaling law for neural language models.
\bibliography{10_reference}

\appendix

\section{Detailed Discussion}
In recent years, deep learning has achieved remarkable progress across a wide range of domains \cite{XiaoYLCH25,chen2021multi}, with large language models (LLMs) demonstrating exceptional capabilities in natural language understanding, reasoning, and complex problem-solving \cite{XiaoYCLLRH25,chen2022ba,chen2024pareto}. The widespread adoption of LLMs has significantly advanced tasks such as automated reasoning and knowledge-intensive applications \cite{chen2024learning,chen2021pareto,chen2021deep}. However, enhancing the mathematical reasoning abilities of LLMs remains a key challenge and an active area of research.

Recent advancements in PRMs have demonstrated their potential to enhance mathematical reasoning in LLMs. Notably, ~\citet{zhang2025lessons} and \citet{lightman2023lets} have identified challenges related to data annotation and evaluation methods, with Zhang et al. proposing a consensus filtering mechanism to improve performance. Meanwhile, ~\citet{ma2024learning} developed a heuristic greedy search algorithm during inference, surpassing the Chain of Thought technique on mathematical benchmarks. ~\citet{zhang2024entropy} also introduced an entropy-regularized PRM that leverages KL-regularized Markov Decision Processes to achieve significant improvements on MATH and GSM8K benchmarks. Additionally, Lightman et al. found that process supervision outperforms outcome supervision, solving 78\% of problems in their test set and demonstrating the benefits of active learning.

Further research into PRMs has also highlighted the development of automated models like Math-Shepherd~\citep{wang2024math}, which minimize manual annotations and demonstrate notable improvements in LLM performance on mathematical tasks.~\citet{setlur2024rewarding} introduced Process Advantage Verifiers (PAVs) that optimize reasoning steps and increase efficiency in search and reinforcement learning. These studies collectively emphasize the role of PRMs in advancing LLMs' ability to reason mathematically, providing a foundation for future research in this field.

\textbf{Test-Time Scaling}
Test-time scaling strategies have emerged as crucial for optimizing the performance of LLMs. ~\citet{setlur2024rewarding} illustrated that the efficient allocation of test-time compute could outperform larger models, achieving over a fourfold increase in efficiency. The work of ~\citet{chen2024simple} proposed a two-stage algorithm with scaling laws, which show exponential reductions in failure rates as test-time compute increases. ~\citet{zhang2024scaling} also explored the interplay of scaling factors in LLM finetuning, discovering that model scaling yields greater benefits than data scaling. Meanwhile, ~\citet{yue2024inference} highlighted performance gains in retrieval augmented generation (RAG) through optimal inference scaling.

These strategies underscore the importance of optimizing test-time computation through methods like knockout tournaments and iterative prompting, illustrating substantial potential to enhance LLM performance across different tasks and conditions.

\textbf{Scaling Laws for Reward Modeling}
Research in scaling laws within reinforcement learning (RL) and reward modeling has identified consistent power-law relationships\cite{wang2024scaling}. ~\citet{gao2023scaling} revealed distinctions between optimization methods for proxy and gold reward models, while ~\citet{neumann2024alphazero} noted that player strength in multi-agent RL scales with neural network parameters and compute. ~\citet{rafailov2024scaling} explored overoptimization in Direct Alignment Algorithms and found degradation patterns similar to classic RLHF when KL budgets increase. ~\citet{hilton2023scaling} examined 'intrinsic performance', establishing power-law relationships in single-agent RL concerning model size and environment interactions.

These findings illustrate the scalability of machine learning models and suggest that, while larger models enhance sample-efficiency, overoptimization against imperfect reward models may hinder true performance, aligning with Goodhart's law. Understanding these scaling laws is essential for AI alignment, model design, and optimizing compute-efficiency.

\section{Detailed Discussion of Process Reward Modeling}

In recent years, Process Reward Models (PRMs) have emerged as a groundbreaking approach to enhancing the reasoning capabilities of Large Language Models (LLMs). This section provides a comprehensive overview of PRMs, detailing their fundamental concepts, training methodologies, and practical applications.

\subsection{Model Architecture}

The PRMs are initialized from the Qwen2.5 series models, with the original language modeling head replaced by a scalar-value head designed for evaluating reasoning steps. This head consists of two linear layers that output a confidence score for each step. The architecture is tailored to assess both the correctness and the utility of each reasoning step within a solution.

\subsection{Optimization Process}

Training the PRM involves optimizing the network parameters to minimize the discrepancy between predicted correctness scores and the target soft labels.

1. Loss Function: The primary objective is to minimize a cross-entropy loss, which measures how well the predicted probabilities align with the soft labels. This loss function inherently accommodates the probabilistic nature of the labels generated via Monte Carlo simulations.

2. Optimization Algorithm: We employ the AdamW optimizer, a variant of the Adam optimization algorithm that incorporates weight decay, to enhance generalization and convergence. Key hyperparameters such as learning rate, batch size, and weight decay are tuned based on validation performance.

3. Regularization Techniques: To prevent overfitting and ensure robustness, regularization methods such as dropout and early stopping are applied. Dropout introduces randomness during training by temporarily removing units from the network, while early stopping halts training when performance on a validation set ceases to improve.

4. Training Iterations: The training process is conducted over multiple epochs, with periodic evaluation on a held-out validation set to monitor progress and adjust hyperparameters as necessary.

Through this comprehensive training process, the PRM learns to accurately predict the correctness of intermediate steps in reasoning tasks. This capability forms the basis for leveraging PRM insights during test time, enabling more effective search and optimization strategies that enhance the LLM's problem-solving performance.

\subsection{Integration into LLMs}

Once trained, PRMs are integrated with LLMs to provide real-time feedback during the reasoning process. At each step, the LLM generates potential actions (next reasoning steps), which are evaluated by the PRM. The feedback is used to adjust the policy, guiding the model towards more accurate reasoning pathways. This iterative refinement leads to improved performance in complex problem-solving tasks, as demonstrated in our experiments.

By establishing a robust process reward modeling system, we lay the foundation for incorporating detailed and actionable feedback into LLM-based reasoning. This preliminary step is vital for the overall effectiveness of the Test-Time Scaling framework, setting the stage for advanced capabilities in LLM reasoning through reinforcement learning and guided search.

\subsection{Search Strategies}

1. \textbf{Best-of-N Sampling.}
   Generation: Generate \( N \) candidate solutions for a given problem.
   Evaluation: Each candidate solution is evaluated using the PRM, which assigns a reward \( R_i \) to each solution \( i \).
   Selection: The solution with the highest reward is selected:
     $$
     \text{Best Solution} = \arg\max_{i \in \{1, 2, \ldots, N\}} R_i
     $$
   This method increases the likelihood of finding a correct solution by exploring diverse reasoning paths.

2. \textbf{Beam Search.}
   Initialization: Start with an initial set of paths (beam width \( K \)).
   Expansion: At each step, expand each path by generating possible next steps.
   Evaluation: Use the PRM to score each expanded path.
   Selection: Retain the top \( K \) paths with the highest cumulative scores:
     $$
     \text{Top Paths} = \text{Select Top } K \text{ from } \{ \sum_{t=1}^{T} R(x_t) \}
     $$
   This approach balances exploration and exploitation, focusing computational resources on the most promising paths.

3. \textbf{Monte Carlo Tree Search (MCTS).}
   Tree Structure: Represent the reasoning process as a tree where nodes are states and edges are actions.
   Selection: Use a policy, such as Upper Confidence Bound for Trees (UCT), to select the next node:
     $$
     \text{UCT}(s, a) = Q(s, a) + c \cdot \sqrt{\frac{\ln N(s)}{N(s, a)}}
     $$
     where \( Q(s, a) \) is the estimated value of action \( a \) from state \( s \), \( N(s) \) is the visit count of state \( s \), and \( N(s, a) \) is the visit count of action \( a \) from state \( s \).
   Simulation: Perform rollouts to simulate the outcome of following a particular path.
   Backpropagation: Update the value estimates of nodes based on the results of the rollouts.
   MCTS effectively balances exploration and exploitation, allowing the model to explore vast solution spaces strategically.

4. \textbf{Majority Voting.}
   Generation: Generate multiple candidate solutions.
   Evaluation: Evaluate each solution using the PRM.
   Aggregation: Aggregate the final answers, selecting the most frequently occurring answer:
     $$
     \text{Final Answer} = \text{Mode}(\{A_1, A_2, \ldots, A_N\})
     $$
   Majority voting leverages the collective insights from multiple solutions, enhancing the robustness of the final answer.

\section{Decoding Phase}
\label{sec:appendix}

During test-time inference, the PRM guides the decoding process by providing real-time feedback on the correctness of each reasoning step. The decoding process involves:

1. Initial Step Generation:
Start with the problem statement \( Q \) and generate the initial reasoning step \( x_1 \).
Evaluate \( x_1 \) using the PRM to obtain a correctness score \( p_1 \).

2. Iterative Step Evaluation:
For each subsequent step \( x_t \), generate candidate steps \( x_{t+1} \) based on the current reasoning context.
Evaluate each candidate step using the PRM, selecting the one with the highest score.
Update the reasoning sequence and continue until the final answer is reached.

3. Score Aggregation:
Aggregate the scores of individual steps to compute the overall score for each candidate solution.
Use strategies such as PRM-Last (considering the score of the last step) or PRM-Min (considering the minimum score across steps) to determine the final solution.

\section{Computational Considerations}

Efficient test-time scaling requires careful management of computational resources. Key considerations include:

1. Parallelization:
Parallelize the generation and evaluation of candidate solutions to speed up the inference process.
Utilize modern hardware accelerators, such as GPUs, to handle the increased computational load.

2. Resource Allocation:
Dynamically allocate computational resources based on the complexity of the problem and the desired accuracy.
Implement adaptive strategies to balance the depth and breadth of search, optimizing resource usage.

3. Caching and Pruning:
Cache intermediate results to avoid redundant computations during iterative evaluations.
Prune low-scoring paths early in the search process to focus computational efforts on promising candidates.

Implementing test-time scaling for PRMs requires careful consideration of computational resources and efficiency. The generation and evaluation of multiple candidate solutions can be computationally intensive, necessitating optimized algorithms and parallel processing techniques.

\section{Benefits of Test-Time Scaling With PRM}

Implementing test-time scaling with PRMs offers several advantages:

1. Efficiency: By allocating compute resources dynamically, test-time scaling reduces unnecessary computation for simpler tasks, leading to more efficient use of available resources.

2. Improved Accuracy: The PRM's step-wise feedback enables the model to correct errors and refine solutions iteratively, resulting in higher accuracy and more reliable outputs.

3. Flexibility: Test-time scaling provides a flexible approach to model deployment, allowing for adjustments based on task complexity and available computational resources. This adaptability is particularly valuable in real-world applications where resource constraints may vary.

4. Cost-Effectiveness: By optimizing the use of test-time compute, organizations can achieve significant cost savings, as fewer resources are required to achieve high-quality results.

\begin{table*}[ht]
\centering
\begin{tabular}{lccc}
\hline
\textbf{Dataset} & \textbf{Total Size} & \textbf{Training Size} & \textbf{Description} \\
\hline
PRM800k & 800k & 400k & Original mathematical reasoning dataset \\
Math-Shepherd & 445k & 400k & Mathematical reasoning with step-level annotations \\
ASLAF (Ours) & 1.2M & 400k & Combined and filtered dataset using our method \\
\hline
\end{tabular}
\caption{Comparison of datasets used for PRM training. For fair comparison, we sampled the same number of examples (400k) from each dataset for training.}
\label{tab:datasets}
\end{table*}

\begin{figure*}[h!]
    \centering
    \vspace{-0.15in}
    \begin{minipage}{0.24\linewidth} 
        \centering
        \includegraphics[width=\linewidth]{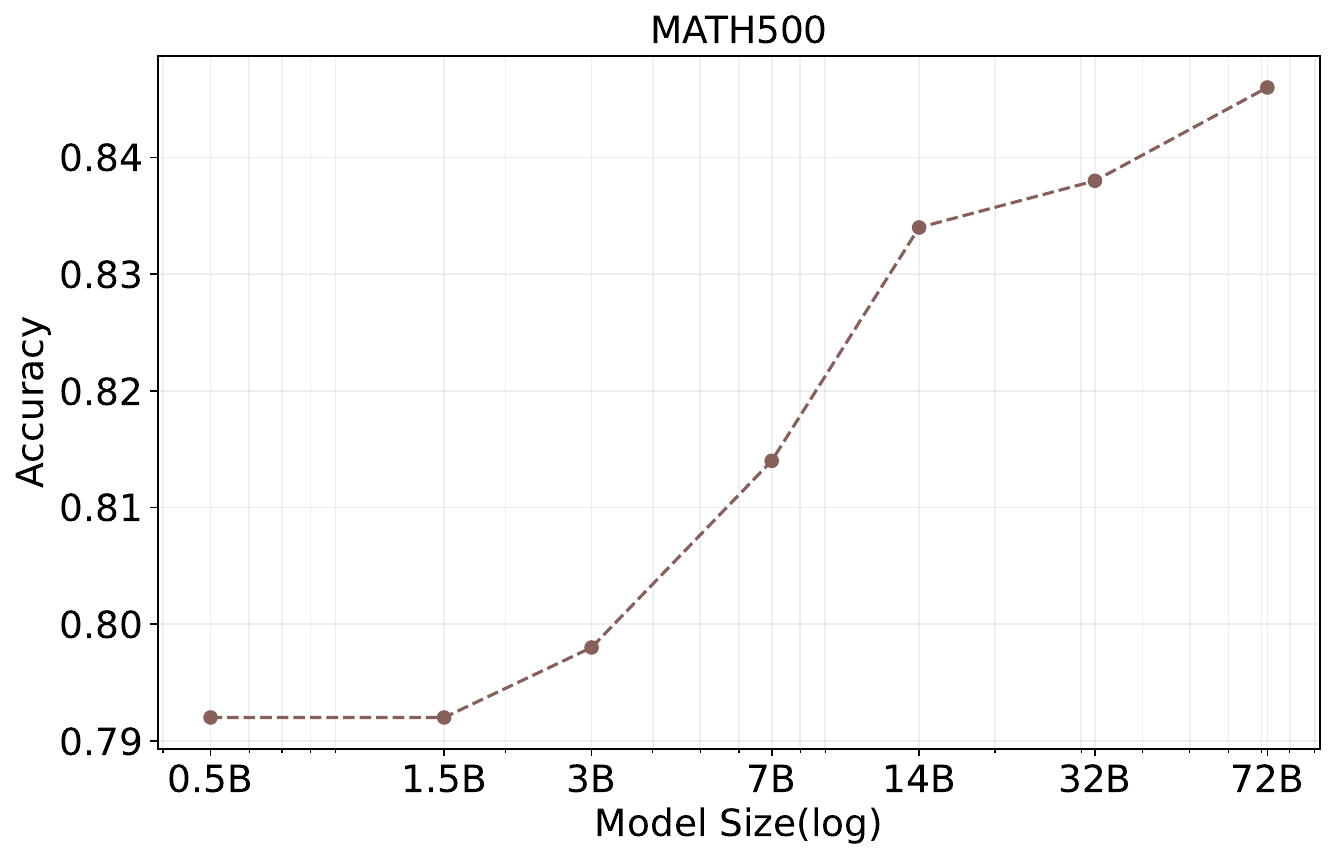}
        \label{fig:training_loss_tokens_lr_v3}
    \end{minipage} 
    \hfill
    \begin{minipage}{0.24\linewidth} 
        \centering
        \includegraphics[width=\linewidth]{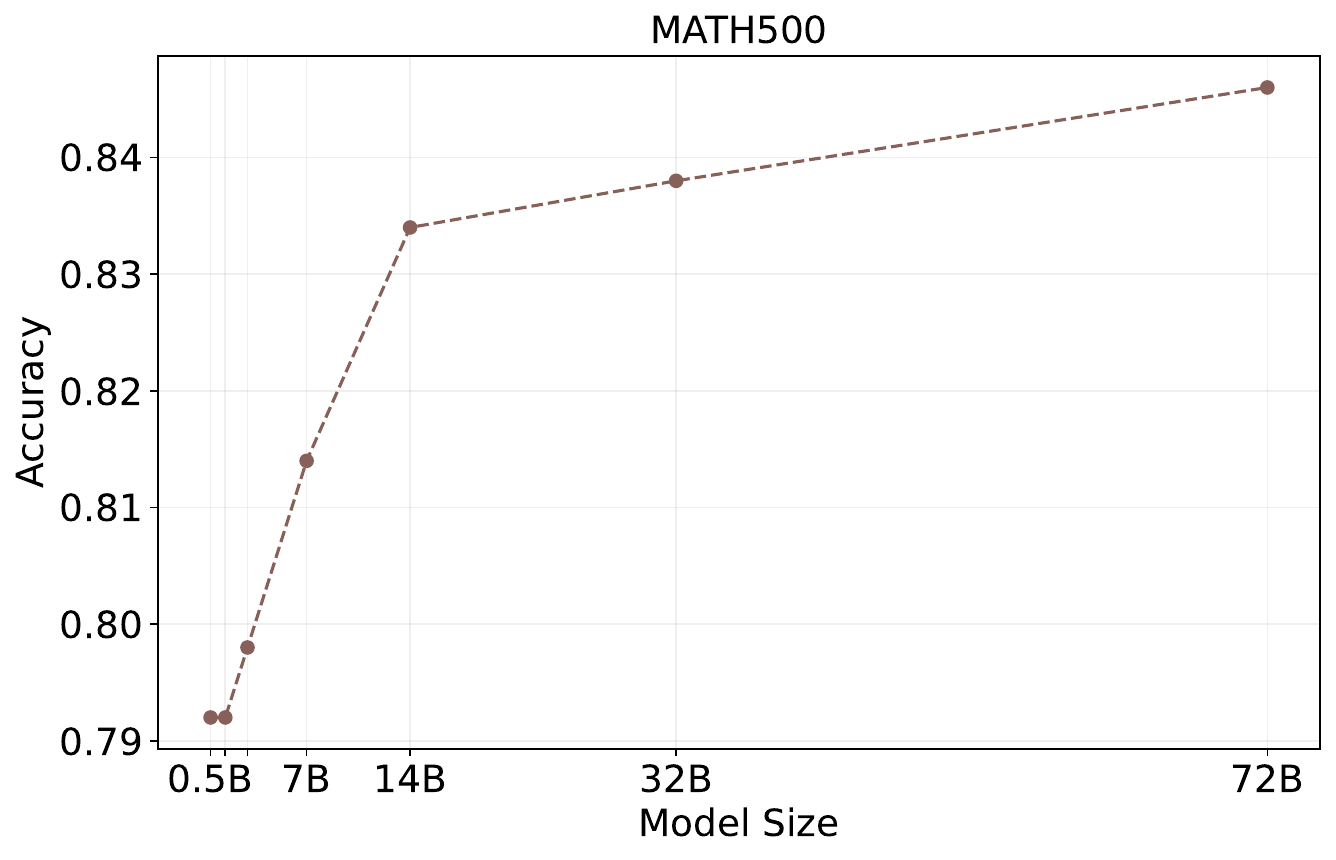}
        \label{fig:training_loss_tokens_lr_v3}
    \end{minipage}  
    \hfill
    \begin{minipage}{0.24\linewidth} 
        \centering
        \includegraphics[width=\linewidth]{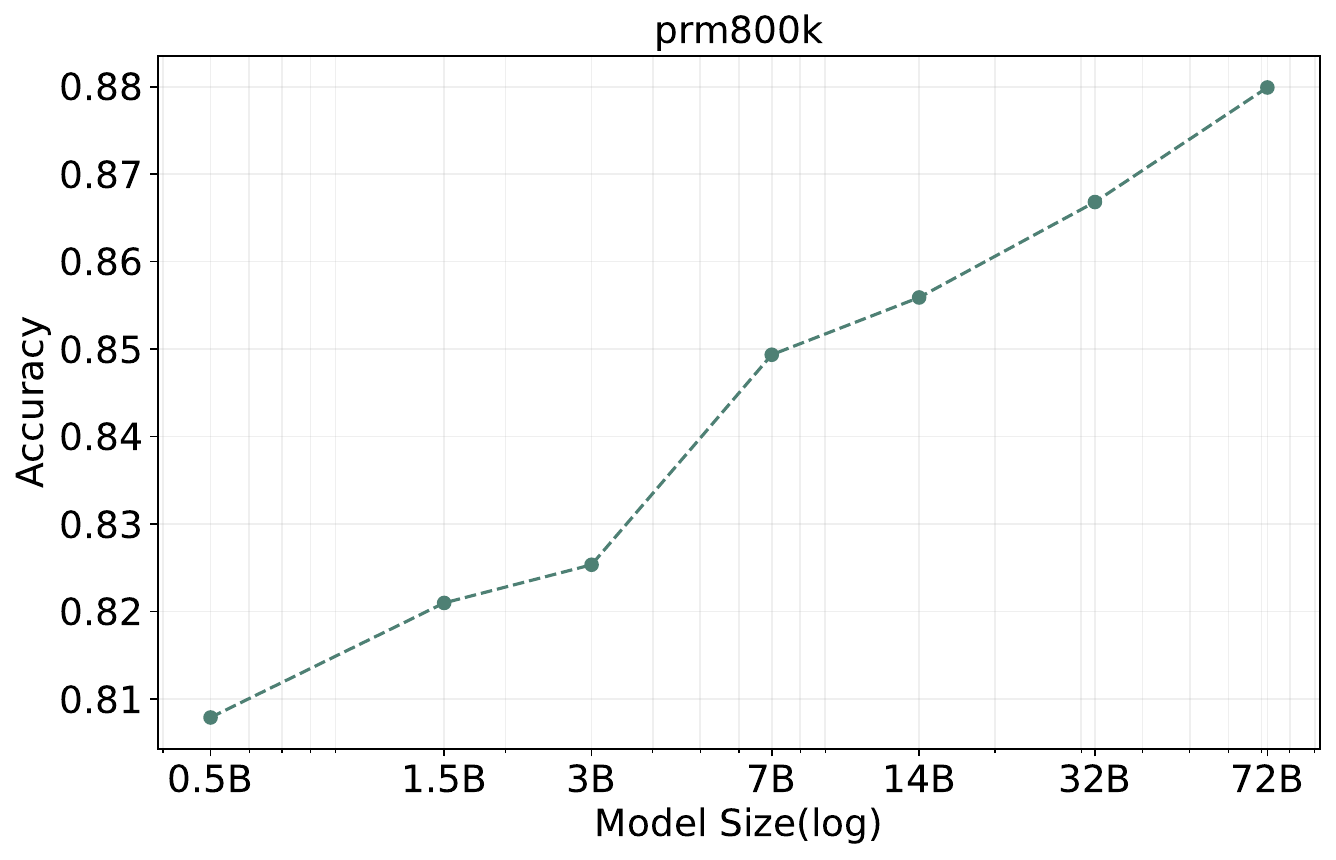}
         \label{fig:training_loss_steps_1B}
    \end{minipage}  
    \hfill
    \begin{minipage}{0.24\linewidth} 
        \centering
        \includegraphics[width=\linewidth]{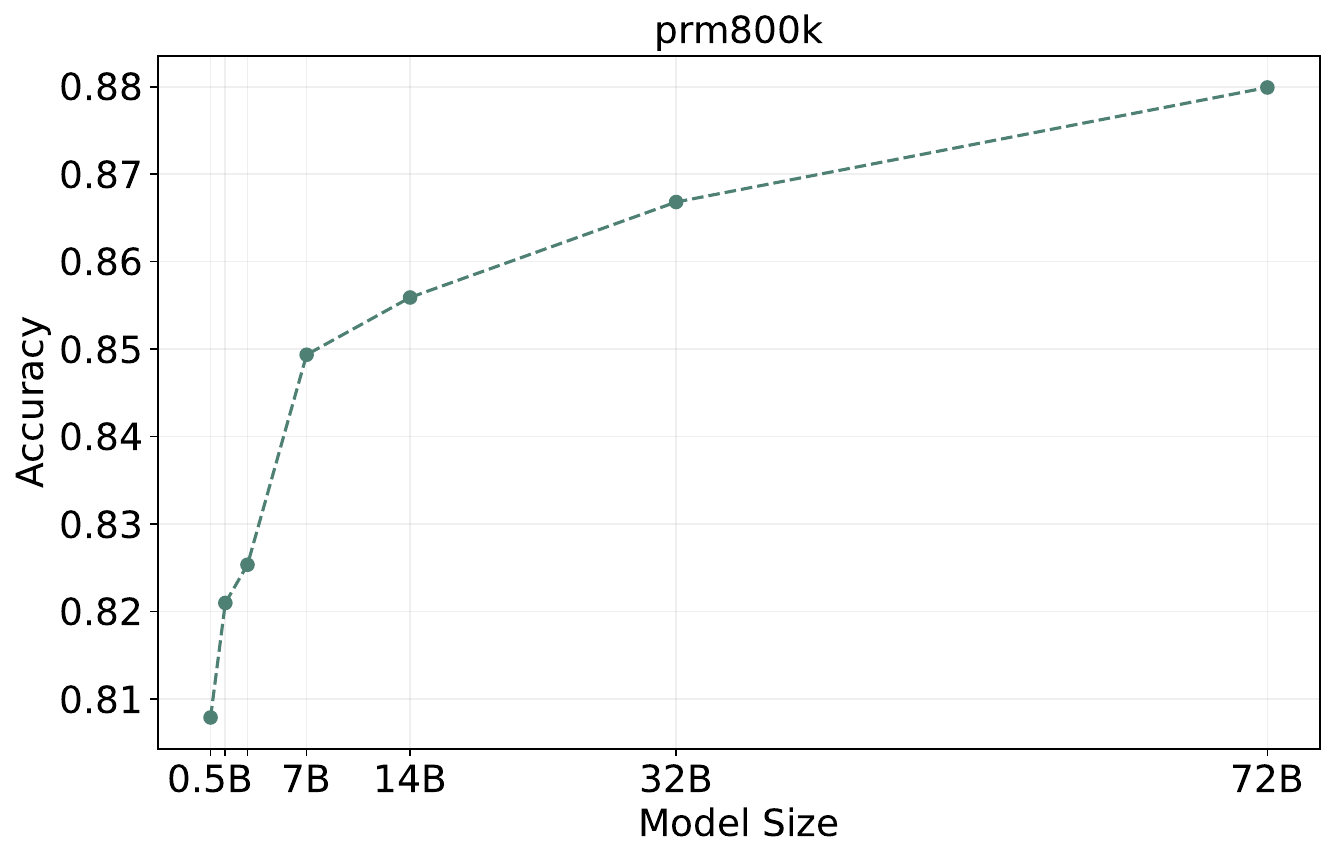}
         \label{fig:training_loss_steps_1B}
    \end{minipage}  
    \vspace{-0.25in}
    \caption{Comparison of the performance of PRMs with varying model sizes during inference using the same language model. The results demonstrate that as model size increases, accuracy improves rapidly, indicating a better capability to capture the complexity necessary for solving mathematical problems.}
    \label{fig:PRM_Models}
\end{figure*}

% \section{Categorization of LiveCodeBench}\label{app:Cate_live}
% We use Qwen2.5-72B to classify the categories of code problems, where the prompt is 

% f"Here is a question for you to code, please give a tag about what kind of algorithm is it?\verb|\|nMath? Graph? Data Structure? Dynamic Planning?\verb|\|n\verb|\|n\{\textit{problem}\}\verb|\|n\verb|\|n1. Math 2. Graph 3. Data Structure 4. Dynamic Planning\verb|\|nAnswer (in 1, 2, 3, 4): "    

\begin{figure*}[t]
    \centering
    \small
\begin{multiqa}
    \begin{questionlayer}
You are given a string s and an integer k. Encrypt the string using the following algorithm:
For each character c in s, replace c with the $k^{th}$ character after c in the string (in a cyclic manner).
Return the encrypted string.\\
...

Constraints:

$1 <= s.length <= 100$, 
$1 <= k <= 10^4$, 
s consists only of lowercase English letters.
    \end{questionlayer}
%     \begin{questionlayer}
    \begin{reasonlayer}

\textbf{Step 1}: \begin{verbatim}class Solution:
    def getEncryptedString(self, s: str, k: int) -> str:
        encrypted_string = ""
        length = len(s)
        for char in s:
            # Find the k-th character after 'char' in a cyclic manner 
            new_char = s[(s.index(char) + k) % length]
            encrypted_string += new_char
        return encrypted_string\end{verbatim}
\begin{flushright}
\textcolor{red}{$0.5371$}
\end{flushright}

\textbf{Step 2}:

However, the above solution uses `s.index(char)` which will not work correctly if there are duplicate characters in the string `s`. Instead, we should use the current index of the character in the loop. Here is the corrected solution:
\begin{flushright}
\textcolor{red}{$0.7563$}
\end{flushright}

\textbf{Step 3}: \begin{verbatim}class Solution:
    def getEncryptedString(self, s: str, k: int) -> str:
        encrypted_string = ""
        length = len(s)
        for i in range(length):
            # Find the k-th character after the current character in a cyclic manner
            new_char = s[(i + k) % length]
            encrypted_string += new_char
        return encrypted_string\end{verbatim}
\begin{flushright}
        \textcolor{red}{$0.875$}
\end{flushright}
    % \textbf{Step 1}: To find the remainder of 2004 divided by 12, we can use the division algorithm, which states that for any integers \(a\) and \(b\) (with \(b > 0\)), there exist unique integers \(q\) (the quotient) and \(r\) (the remainder) such that \(a = bq + r\) and \(0 \leq r < b\). \textcolor{red}{$[0.958, 0.988]$} \\

    % \textbf{Step 2}: Here, \(a = 2004\) and \(b = 12\). We need to find \(q\) and \(r\) such that:
    % \[
    % 2004 = 12q + r
    % \]
    % where \(0 \leq r < 12\). \textcolor{red}{$[0.924, 0.987]$} \\

    % \textbf{Step 3}: First, we perform the division of 2004 by 12:
    % \[
    % 2004 \div 12 = 167
    % \]
    % This means the quotient \(q\) is 167.
    %  Now, we can find the remainder \(r\) by multiplying the quotient by the divisor and subtracting the result from the original number:
    % \[
    % 2004 - 12 \times 167 = 2004 - 2004 = 0
    % \]
    % \textcolor{red}{$[0.777, 0.719]$} \\

    % \textbf{Step 4}: Thus, the remainder \(r\) is:
    % \[
    % r = 0
    % \]
    
    % \textbf{Step 5}: Therefore, the remainder when 2004 is divided by 12 is:
    % $\boxed{0}$. \textcolor{red}{$[0.622, 0.349]$}

    \end{reasonlayer}
    % \begin{answerlayer}
    % An integer is divisible by 12 if and only if it is divisible by both 3 and 4. 

    % Because \( 2 + 0 + 0 + 4 = 6 \) is divisible by 3, 2004 is divisible by 3. 
    
    % Also, the last two digits of 2004 form a multiple of 4, so 2004 is divisible by 4 as well. 
    
    % Therefore, 2004 is divisible by 12 and hence leaves a remainder of \( \boxed{0} \) when divided by 12.
    
    % % The remainder is $\boxed{0}$.
    % \end{answerlayer}

\end{multiqa}
    \caption{PRM score comparison for a case with a correct answer.}\label{fig-ex2}
\end{figure*}

\end{document}